\documentclass[journal]{IEEEtran}
\usepackage{cite}
\usepackage{amsmath,amssymb,amsfonts}
\usepackage{algorithm}
\usepackage{algpseudocode}
\usepackage{graphicx}
\usepackage{textcomp}
\usepackage{xcolor}
\usepackage{booktabs}
\usepackage{multirow}
\usepackage{caption}
\usepackage{float}
\usepackage{paralist}
\usepackage{tabularx}
\usepackage{threeparttable}
\usepackage{enumitem}
\usepackage{booktabs}
\usepackage{multirow}
\usepackage{threeparttable}
\usepackage{array}
\usepackage{hyperref}
\usepackage{etoc}
\usepackage{subcaption}

\ifCLASSINFOpdf

\else

\fi

\usepackage{amsmath}
\usepackage{listings}
\usepackage{minted}

\begin{document}


\def \FNAME{SNNF}

\title{SNNF: An SNN-based Near-Sensor Noise Filter\\for Dynamic Vision Sensors}

\author{{Yahan Yang*,
        Pradeep Kumar Gopalakrishnan*,~\IEEEmembership{Senior Member, IEEE},
        Chang Chip Hong,~\IEEEmembership{Fellow,~IEEE},
        Arindam Basu,~\IEEEmembership{Fellow,~IEEE}}
        
\thanks{This work was supported by the Research Grants Council of the HK SAR, China (Project No. CityU 11208125). 

*Yahan Yang and Pradeep Kumar Gopalakrishnan contributed equally to this work. (Corresponding author:  Arindam Basu. e-mail:  arinbasu@cityu.edu.hk.)}
}



\maketitle

\maketitle

\begin{abstract}
Dynamic Vision Sensors (DVS) exhibit exceptional dynamic range and low power consumption, making them ideal for edge applications in the Internet of Video Things (IoVT). However, their output is often degraded by spurious Background Activity (BA) noise, leading to unnecessary computational overhead. This paper proposes SNNF, a near-sensor BA noise filter that integrates a compact Event-Based Binary Image (EBBI) representation, a parallel memory architecture, and a single-layer Spiking Neural Network (SNN) classifier. Trained on representative DVS data, the SNN distinguishes signal events from noise with an AUC of 0.89 on standard datasets. The binary-array-based EBBI eliminates timestamp dependency, significantly reducing memory footprint. Moreover, the SNN’s spike-based computation replaces power-hungry multipliers with simple accumulation logic and minimizes inter-neuron data width, resulting in an extremely hardware-efficient design. 
FPGA implementation results show that SNNF reduces memory and logic resources to approximately 11\% and 40\%, respectively of state-of-the-art filters \cite{Rios-Navarro2023}, while achieving a throughput of 29 Mega events per second (Meps). In a 65 nm CMOS ASIC implementation, SNNF achieves 44.4 Meps with an area and power consumption of only $\sim 13\%$ and $\mathord{<}5\%$ of the corresponding ANN-based designs. These results demonstrate that SNNF provides an excellent balance between filtering accuracy and hardware efficiency, making it highly suitable for resource-constrained, near-sensor deployment.
\end{abstract}


\IEEEpeerreviewmaketitle

\addtocontents{toc}{\protect\setcounter{tocdepth}{-10}}

\section{Introduction}

\IEEEPARstart{T}{he} Dynamic Vision Sensor (DVS), also known as an event-based or neuromorphic vision sensor, is a bio-inspired imaging device that operates asynchronously and independently at the pixel level. Each pixel generates an event only when the temporal contrast of incident light exceeds a predefined threshold, leading to a sparse and efficient data encoding ideal for low-power applications \cite{Delbruck2008}. Under static conditions with constant illumination, an ideal DVS produces no output, enabling effective background suppression, especially in stationary camera setups. This data reduction capability, combined with a high dynamic range ($>120$ dB), high temporal resolution, low latency, and kHz-level effective frame rates, makes the DVS highly suitable for a wide range of applications such as autonomous driving \cite{Maqueda2018}, drone navigation \cite{Gallego2022}, intelligent transportation systems \cite{Acharya2019, Ussa2023}, high speed and high dynamic range (HDR) imaging \cite{Rebecq2019, Rebecq2021}, motion estimation \cite{Benosman2012, Benosman2014}, video reconstruction \cite{Nie2021}, robotic tracking \cite{Delbruck2013, Ramesh2021}, human-machine interaction \cite{Amir2017, Tapiador2020}, depth estimation \cite{devulapally2024multi}, and optical flow estimation \cite{zhu2018ev}.

A significant challenge for DVS-based systems is Background Activity (BA) noise. Unlike signal events, BA noise is caused not by variations in external light intensity, but by intrinsic physical imperfections within the sensor's pixel circuitry, such as junction leakage currents and temporal noise \cite{Lichtsteiner2008, Liu2015}. This noise manifests quite differently under bright and dark conditions, as it is dominated by distinct noise sources \cite{tian2001analysis, Guo2022}. It becomes particularly prominent under low-light conditions \cite{Lichtsteiner2008, Hu2021}, increasing the computational burden and energy consumption of downstream processing. For always-on edge vision systems operating under strict power constraints—-typically several tens of milliwatts \cite{Gallego2022}—-false events can trigger unnecessary system activations, severely reducing energy efficiency \cite{Khodamoradi2021}. Consequently, effective near-sensor filtering is essential. 

Near-sensor computing can effectively reduce data transfer between the sensor and the computational platform, thereby conserving bandwidth and lowering overall energy consumption \cite{zhou2020near}. A near-sensor filter classifies incoming events as signal or noise, transferring only relevant data to subsequent stages. 
However, as sensor resolutions increase from $128 \times 128$ to $1280 \times 960$ pixels \cite{Gallego2022}, the memory requirement and energy consumption of traditional filters rise considerably. For instance, the Spatio-Temporal Correlation Filter (STCF) \cite{Guo2022} and the Multi-Layer Perceptron Filter (MLPF) \cite{Rios-Navarro2023} both exhibit significant growth in storage resources as resolution scales. STCF utilizes Surface of Active Events (SAE) for event data representation, but it does not fully exploit spatio-temporal characteristics as only the most recent events within a local neighborhood are counted. While MLPF uses Time Surface (TS) for data storage and a deep neural network for denoising, it does not adopt a Spiking Neural Network (SNN) architecture, which fails to leverage the inherent temporal sparsity of DVS data.

To address the need for high-performance and hardware-efficient filtering, we propose SNNF—a novel near-sensor BA noise filter. SNNF leverages an SNN for classification, supported by a compact Event-Based Binary Image (EBBI) representation \cite{Acharya2019}, \cite{Mohan2022} and a parallel memory bank architecture. SNNF enables high-accuracy discrimination between noise and signal events after training on representative DVS data samples. Key advantages of SNNF include: (1) state-of-the-art denoising performance on standard datasets; (2) significant reductions in memory usage by employing binary representation and cyclic reset instead of timestamp storage; (3) high-throughput event processing via parallel memory access; and (4) a minimal resource footprint achieved by a single hidden-layer SNN with only $\approx$30 hidden neurons. FPGA and ASIC implementation results confirm that SNNF significantly reduces resource utilization compared to existing high-performance filters, making it ideal for resource- and power-limited edge applications. To the best of our knowledge, SNNF is the first filter to combine 1-bit binary representations with a hardware-accelerated SNN to provide a scalable solution for high-resolution, low-power DVS noise suppression.


\section{Background and Related Works}
\subsection{Data Structures}
The DVS data flow is based on events. Each event is represented by a quadruplet $e(x, y, t, p)$ consisting of:
\begin{compactitem}
\item $x$ - Column number of the event-generating pixel
\item $y$ - Row number of the event-generating pixel
\item $t$ - An $n_T$-bit event-timestamp (typically $32$-bit) 
\item $p$ - Polarity ($1/0$: signifying an increase/decrease in illumination)
\end{compactitem}

The choice of data structures and event representations for event storage and search in BA filters directly impacts hardware implementation cost, computational efficiency, filtering accuracy, and workload scalability.
\subsubsection{Surface of Active Events (SAE)}
The SAE \cite{mueggler2017fast} is an asynchronously updated 2D array $\Sigma_e$ that tracks the arrival time of the most recent event for every pixel. For each incoming event $e_i$, the surface is updated as $\Sigma_e(x,y,p) \leftarrow t_i$. 

\subsubsection{Time Surface (TS)} A TS \cite{HOST} is a normalized representation derived from the SAE by applying a temporal decay kernel. It maps timestamps into a range $[0,1]$. The TS value at pixel ($x,y$) for a given polarity $p$ is calulated as:
\begin{equation}
S_p(x,y,t) = e^{-\frac{t-\Sigma_e(x,y,p)}{\tau}}
\end{equation}
where $\tau$ is a temporal decay constant that determines the ``memory’’ length of the surface. 

Both SAE and TS use \textit{$n_{T}$} bits per entry. The high-bit-depth memory ($n_{T} = 16-32$) to store these surfaces pose a significant bottleneck for near sensor ASIC implementations. 

\subsubsection{Event Based Binary Image (EBBI)} The EBBI \cite{Acharya2019}, \cite{Mohan2022} is a fundamental 2-D representation derived from event streams. 
It uses a binary map where each pixel indicates the presence/absence of an event within an accumulation window. EBBI significantly minimizes storage and bandwidth requirements per pixel to just 1 bit, dramatically lowering the footprint for ASIC implementation, but it loses precious temporal information.

\subsubsection{Bloom Filter (BF)}

The BF \cite{Bloom1970} is a space-efficient probabilistic data structure comprising a bit array of length \textit{N} and \textit{d} independent hash functions. Elements are mapped to \textit{d} positions in the array and set to 1; a query returns ``definitely absent’’ if any bit is 0 or ``possibly present’’ if all \textit{d} bits are 1. 
While highly memory-efficient, BFs are prone to false positives as the element count increases. Furthermore, because hashing disrupts the spatial lattice, BFs lack inherent spatiotemporal correlation \cite{Guo2020}, limiting their use in tasks requiring local analysis of temporal or sequential patterns.
\subsubsection{Voxel Grid} This 3D representation discretizes event streams into spatiotemporal volumes (Height $\times$ Width $\times$ Time), where the time dimension is divided into discrete bins \cite{Zhu2018, Gehrig2019}. While this enables the use of standard 3D CNNs, it introduces temporal quantization errors and entails high computational overhead due to the density of the resulting tensor.

\subsection{Spatiotemporal Correlation Principle}

The most fundamental distinction between BA noise and genuine events lies in their spatiotemporal correlation. \cite{Liu2015}. Events generated by the motion of real objects are spatiotemporally continuous. In contrast, BA noise occurs irregularly at random times and random pixel locations, exhibiting no correlation with surrounding events. 

This underlying principle is known as the spatiotemporal correlation (STC) principle.

\subsection{BA Noise Filters}
\label{BA_filters}
\subsubsection{Background Activity Filter (BAF)}
The BAF proposed in \cite{Delbruck2008} was the first to utilize the STC principle, significantly reducing computational overhead while maintaining effective filtering. It employs an SAE to store event history. Within a size of $3\times 3$ neighborhood, BAF requires only one supporting event during the preceding time $\tau$ in the neighbourhood (excluding the current pixel) to classify an event as a signal. However, this often leads to a high false positive rate and a strong dependency on the hardware interface. Moreover, its memory requirement increases proportional to sensor resolution.

\subsubsection{Spatiotemporal Correlation Filter (STCF)}
STCF \cite{Guo2022} generalizes BAF by requiring $k$ ($k>1$) supporting events and introduces a threshold \textit{s} to guide the signal decision. For a $3\times 3$ neighborhood with \textit{$\tau$} = 10 ms and {$k=4$}, STCF achieves BA noise reduction over a thousand times greater than BAF. However, it does not fully exploit temporal dynamics. While effecive for low-density noise, STCF struggles with high-density BA noise and does not capture complex motion patterns. In complex scenarios, its accuracy may be inferior to algorithms leveraging richer spatiotemporal information, such as those based on multilayer perceptrons (MLPs) or SNNs. Similar to BAF, implementing STCF on hardware requires a large memory array to store the last timestamp of every pixel, which scales linearly with sensor resolution.

\subsubsection{Multilayer Perceptron Denoising Filter (MLPF)}
MLPF \cite{Rios-Navarro2023} represents the event stream using a TS with an 18-bit word width (16 bits for the timestamps; 2 bits for polarity). A $7\times 7$ pixel patch is extracted and fed into a two-layer perceptron to classify events. While MLPF demonstrates superior performance in complex environments, such as driving and surveillance, it does not fully leverage the asynchronous nature of DVS. Additionally, MLPF requires the training and storage of approximately 1,000 weights and biases, and its resource requirements for storing the TS scale significantly with sensor resolution. The reliance on a synchronous artificial neural network (ANN) necessitates power-hungry multipliers and frequent memory access to high-bit-depth timestamps.

\subsubsection{O(N)-Space Spatiotemporal Filter (ONF)}
ONF \cite{Khodamoradi2021} is also based on the STC principle. Unlike STCF, which assigns a fixed memory to every pixel, ONF allocates two 32-bit memory words per row and column to store the latest events. This significantly reduces resource usage. Nevertheless, this design is optimized for sparse event scenarios in dense scenes, its accuracy decreases substantially \cite{Guo2022} due to timestamp overwriting.

\subsubsection{Hashheat Filter} \label{sec:hashheat}

This filter \cite{Guo2020} is a hash-based approach (related to Bloom filter) that operates without storing explicit raw timestamps for every pixel. It leverages locality-sensitive hashing (LSH) and thermal value statistics to discriminate noise, significantly reducing memory consumption. The workflow proceeds as follows: the spatiotemporal attributes (x, y, timestamp) of an incoming event are encoded by \textit{k} independent LSH functions, generating \textit{k} distinct hash values. These values then index a fixed-length array of size \textit{m} (termed the "heat array") to retrieve the corresponding thermal counts. If any of the \textit{k} retrieved thermal values falls below a predefined dynamic threshold, the event is classified as noise. 
A notable limitation is that the heat array requires periodic resetting, which can disrupt the processing of continuous event streams. Furthermore, its performance is suboptimal in scenarios with dense or rapidly changing event patterns.

Existing filters often struggle to simultaneously achieve low resource usage and high denoising accuracy. For instance, efficiency-focused approaches like HashHeat \cite{Guo2020} perform well only under simple motion patterns or low-noise conditions. To address these limitations, we propose SNNF, an efficient near-sensor BA noise filter that introduces two key architectural innovations. First, instead of high-resolution representations like TS, we utilize EBBIs to reduce the memory requirement per pixel to just 1 bit. Second, we utilize an SNN to replace the power-hungry multipliers used in ANN-based filters, such as MLPF, with simple accumulation logic, leveraging the inherent sparsity of DVS data to minimize dynamic power. While the transition from high-resolution timestamps to 1-bit EBBI representations involves a loss of precise temporal information, this is strategically compensated by the SNN's ability to extract complex spatiotemporal correlations. Unlike traditional filters that rely on hard-coded temporal thresholds, the SNN learns to distinguish signal from noise based on the spatial density and structural patterns inherent in the binary frames. This trade-off significantly reduces the memory footprint (by up to $32\times$ per pixel) while maintaining superior filtering accuracy, as the SNN effectively replaces explicit timestamp data with learned feature-based discrimination. 

\subsection{Spiking Neural Networks (SNN)}
SNNs represent a promising approach to hardware-efficient Machine Learning (ML), particularly for edge computing and neuromorphic applications.
Unlike traditional Artificial Neural Networks (ANNs), which use real numbers to represent activations, SNNs are more biologically inspired, mimicking the way real neurons communicate via brief electrical pulses or ``spikes". They operate in an event-based manner, processing information only when a spike occurs.

\begin{figure*}[t]
	\centering
	\vspace{-0.15in}
	\begin{minipage}{1\linewidth}	
		\subfloat[] {
			\includegraphics[width=0.33\textwidth]{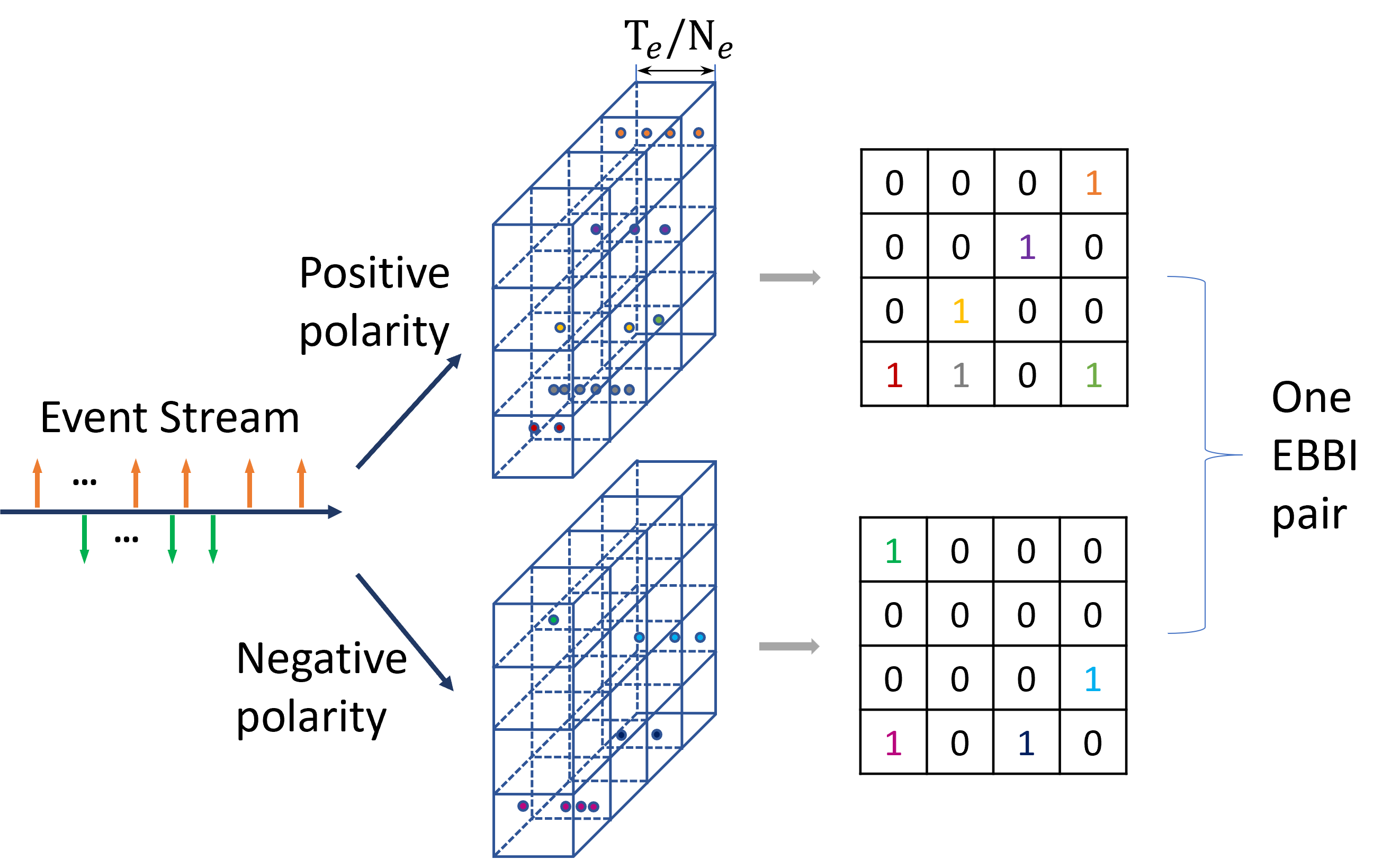}	
		}
        \subfloat[] {
			\includegraphics[width=0.27\linewidth]{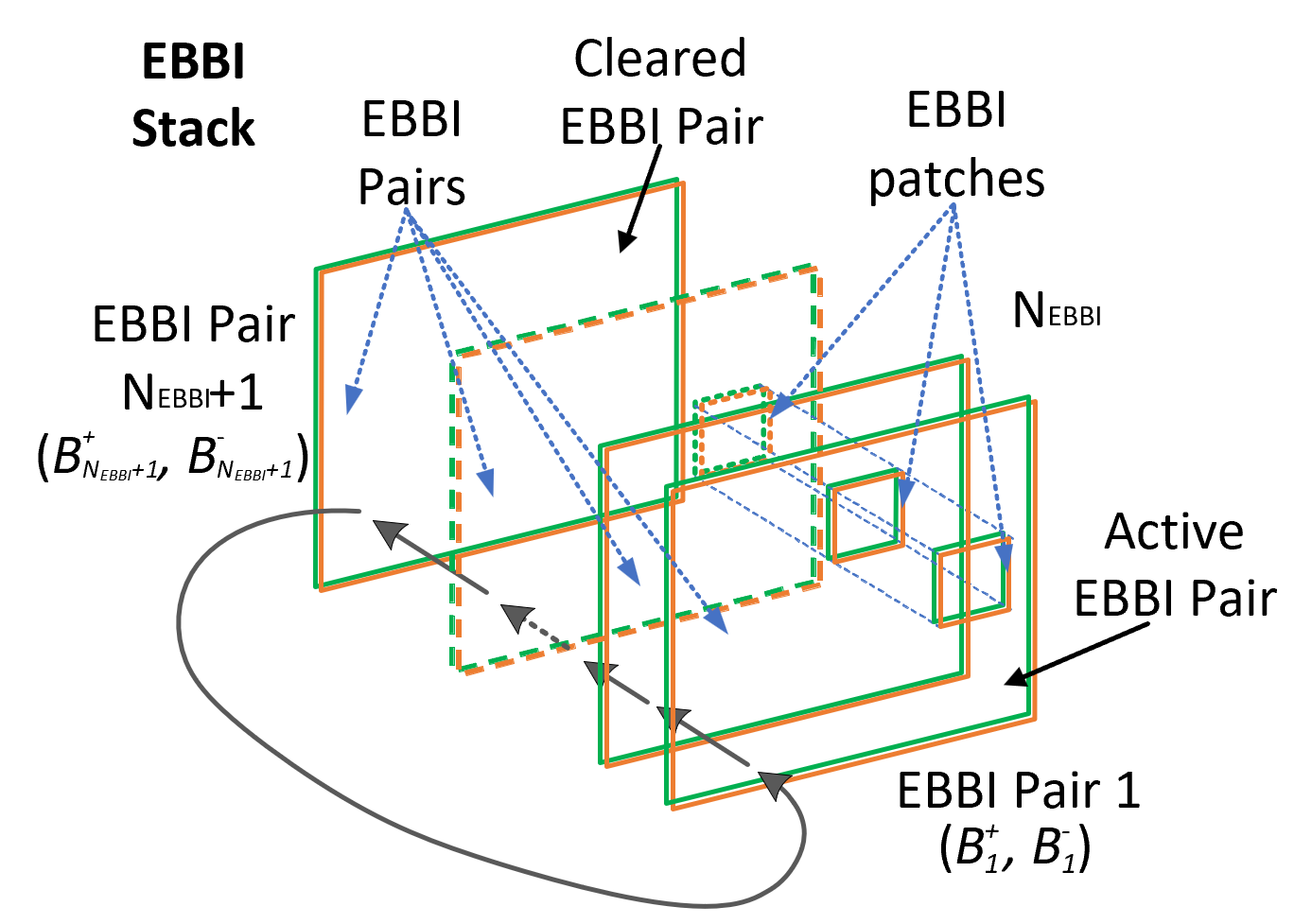}
		}
		\subfloat[] {
			\includegraphics[width=0.38\textwidth]{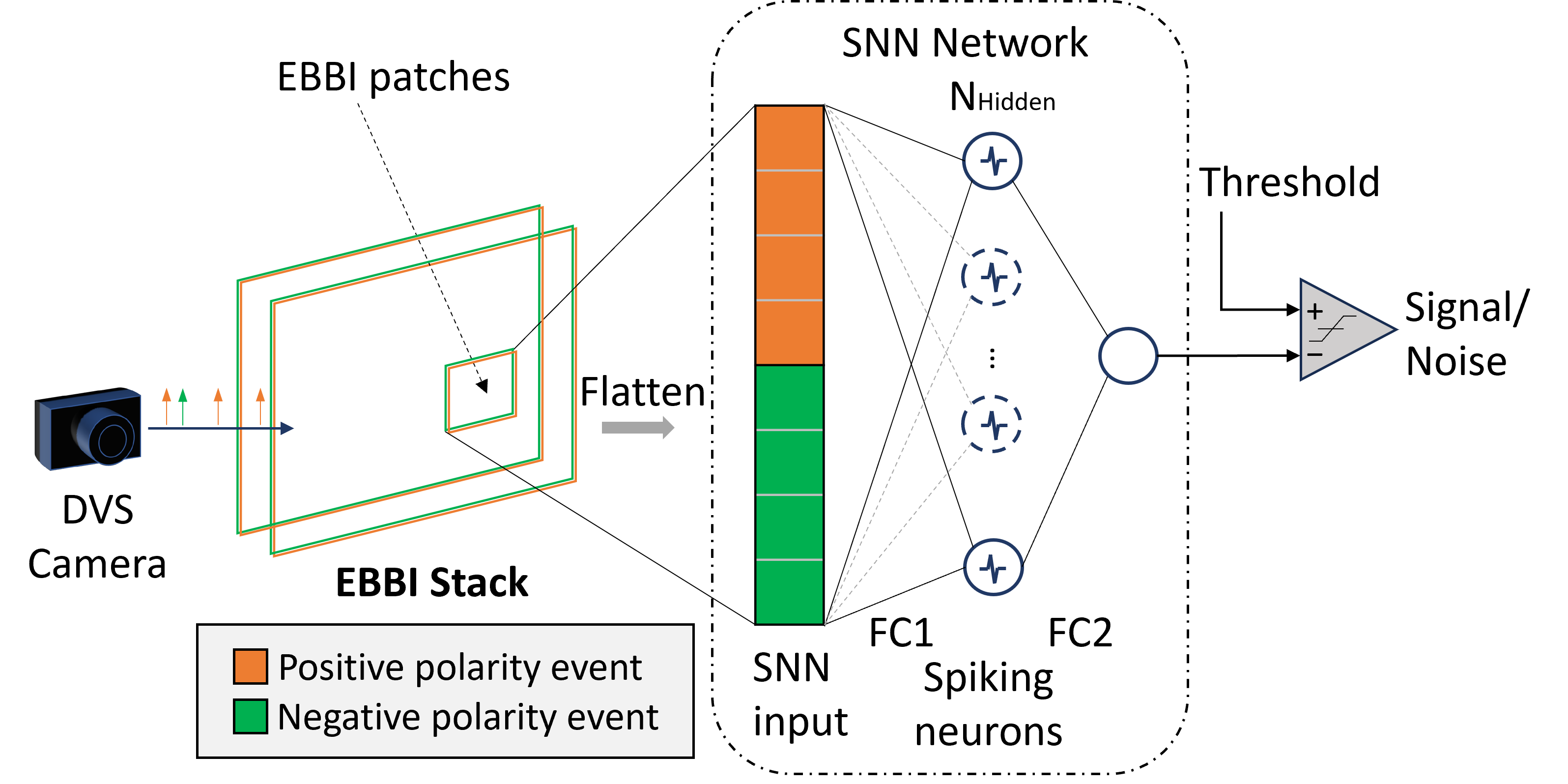}
		}
	\end{minipage}

	\caption{(a) EBBI pair generation from raw events. (b) A total of $N_\emph{EBBI}+1$ EBBI pairs are stacked and accessed in round robin fashion; while the oldest one is being cleared, the remaining $N_\emph{EBBI}$ pairs are used for processing. (c) Overall architecture of the SNNF, where the input event history is stored in the EBBI stack. A patch around the current event is flattened into a vector and passed as input to the SNN, which produces an output indicating whether the current event is signal or noise. }
	\vspace{-0.2in}		
	\label{fig:arc}
\end{figure*}

The core of SNN resides in its spatiotemporal dynamics. A popularly used neuron model is the Leaky Integrate-and-Fire (LIF) neuron, which can be described by the following dynamics:
\begin{align}
\tau_m \frac{dV_m}{dt} &= -V_m(t) + R_m I_{\text{syn}}(t) \notag\\
\text{If}\; V_m(t) &\geq V_\emph{th}, s(t) = 1, V_m(t^+) = V_\emph{reset}, \notag \\
& \text{Else}\; s(t) = 0
\end{align}
where $V_m$, $V_\emph{th}$ and $V_\emph{reset}$ represent the membrane potential, threshold, and reset potentials, respectively. $\tau_m$ is the membrane time constant, which determines the decay rate of the membrane potential, $I_{\text{syn}}(t)$ is the synaptic input current, and $R_m$ is the membrane resistance. 
$s$ indicates whether a spike is generated ($s=1$) or not ($s=0$).

When a neuron receives input pulses, its membrane potential accumulates. Simultaneously, it ``leaks" over time, simulating natural potential decay. Once the potential exceeds $V_\emph{th}$, the neuron ``fires", transmitting a signal to subsequent layers and resetting its own state.

This sparse communication significantly reduces data movement and energy consumption.Furthermore, replacing complex multi-bit activations with binary spikes reduces memory requirements and enables simpler hardware implementations. Specialized neuromophic processors, such as IBM TrueNorth \cite{TrueNorth2015} and Intel Loihi \cite{Loihi2018}, have been designed to exploit these properties for real-time DVS applications, including optical flow computation \cite{haessig2018spiking, Brosch2016}, stereo depth estimation \cite{Dikov2017}, stereo reconstruction \cite{andreopoulos2018low}, object tracking \cite{Glover2019}, etc.

\section{Proposed Work}
Figs. \ref{fig:arc}(a)-(c) depict the operational mechanism of SNNF. The event data input is first converted into a sequence of $N_\emph{EBBI}$ EBBI pairs before being processed by a one-layer fully connected spiking neural network (FCSNN). The network produces one output signal; when this output exceeds a predefined threshold, the event is classified as a signal rather than noise. The hardware architecture of SNNF will be detailed in Section \ref{sec:hardware_arch}.

\subsection{Algorithm}
\label{sec:algorithm}

\subsubsection{EBBI Generation}

EBBI pairs are generated using two distinct methodologies (Fig. \ref{fig:arc}(a)): fixed temporal interval or fixed event count. Each EBBI pair comprises two binary images encoding events with positive and negative polarities, respectively. The spatial dimensions of these images match the sensor resolution of $W\times H$ pixels.

Formally, let the event stream be a sequence $e_i = (x_i, y_i, t_i, p_i)$, where $x_i \in [1, W]$, $y_i \in [1, H]$ are pixel coordinates, $t_i$ is the timestamp, and $p_i \in \{+1, -1\}$ represents the polarity (positive or negative).

\textit{(a) Fixed Time Interval} (duration $T_e$):
The EBBI temporal sequence, $\{B_{{T_e}}^{(0)}, B_{{T_e}}^{(1)}, B_{{T_e}}^{(2)}, \cdots\}$, is defined as:
\begin{equation}
B_{T_e}^{(k)}(x,y,p) = 
\begin{cases} 
1 & \text{if } \exists e_i \in E^{(k)}_{T_e} \text{ s.t. }  (x_i, y_i, p_i)=(x,y,p)  \\
0 & \text{otherwise}
\end{cases}
\end{equation}
where $k \in {0, 1, 2, \cdots}$ is the index of the time window, ${T_e}$ denotes the window duration, and $E^{(k)}_{T_e} = \{e_i | t_i \in [k{T_e}, (k+1){T_e})\}$ is the set of events occuring within the $k$-th temporal window.

\textit{(b) Fixed Event Count} ($N_e$ events per EBBI):
The EBBI temporal sequence, $\{B_{N_e}^{(0)}, B_{N_e}^{(1)}, B_{N_e}^{(2)}, \cdots\}$, is defined as:
\begin{equation}
B_{N_e}^{(m)}(x,y,p) = 
\begin{cases} 
1 & \text{if } \exists e_i \in E^{(m)}_{N_e} \text{ s.t. } (x_i, y_i, p_i) = (x,y,p)  \\
0 & \text{otherwise}
\end{cases}
\end{equation}
where $m \in {0, 1, 2, \cdots}$ is the sequence index, ${N_e}$ is the fixed number of events per set, and $E^{(m)}_{N_e} = \{e_i | i \in [m{N_e}, (m+1){N_e})\}$ represents the $m$-th event set.

\subsubsection{EBBI Stack}

With the arrival of each new event, the corresponding pixel is set in the active EBBI pair, as indicated in Fig. \ref{fig:arc}(b). This continuous update process maintains the active EBBI as a dynamic representation of recent activity. When either the time threshold $T_e$ or event count threshold $N_e$ is reached, the active EBBI is archived as a historical frame in the EBBI stack. Simultaneously, the oldest stored EBBI is cleared and repurposed as the new active buffer. This circular buffering mechanism functions as a First-In-First-Out (FIFO) system, where the oldest EBBI is overwritten to become the newest active buffer. Consequently, the filter maintains exactly ($N_{\text{EBBI}}$+1) EBBI pairs at any given time: one active pair for ongoing event accumulation, ($N_{\text{EBBI}}-1$) historical pairs and one cleared (free) pair ready for reuse. The procedure is detailed in Algorithm \ref{robin}. The EBBI stack is formally defined as $\mathcal{S}$ = $\{(B^{+}_k, B^{-}_k)\}_{k=1}^{N_{\text{EBBI}}+1}$, where $B^{+}_k$ and $B^{-}_k$ are the binary images for positive and negative polarities, respectively at stack position $k$.

\begin{algorithm}
    \caption{EBBI Stack Round-Robin}
    \label{robin}
    \begin{algorithmic}[1] 
    
    \Procedure{Initialization}{}
        \State \textbf{Input:} $N_{\text{EBBI}}$ \Comment{Number of EBBI pairs}
        \For{$k \gets 1$ to $N_{\text{EBBI}}+1$}
            \For{each coordinate $(x, y)$}
                \State $B_k^+(x, y) \gets 0, B_k^-(x, y) \gets 0$
            \EndFor
        \EndFor
        \State $\mathit{ptr}_{\text{active}} \gets 1$ \Comment{Active EBBI pointer}
        \State $\mathit{ptr}_{\text{clear}} \gets N_{\text{EBBI}}+1$ \Comment{oldest EBBI pointer}
        \State $\mathit{event\_count} \gets 0$ \Comment{Current EBBI event counter}
        \State $\mathit{t\_start} \gets t_1$ \Comment{First event timestamp}
    \EndProcedure
\State \Comment{Process incoming event $e_i$ with polarity $p_i \in \{+,-\}$}
    \Procedure{ProcessEvent}{$e_i = (x_i, y_i, t_i, p_i)$}
        \State \textbf{Step 1: Update active EBBI}
        \State $B_{\mathit{ptr}_{\text{active}}}^{p_i}(x_i, y_i) \gets 1$

        \State \textbf{Step 2: Increment event counter}
        \State $\mathit{event\_count} \gets \mathit{event\_count} + 1$

        \State \textbf{Step 3: EBBI transition}
        \If{$(t_i - \mathit{t\_start} \geq T_e)$ \textbf{or} $(\mathit{event\_count} \geq N_e)$}

        \State $\mathit{ptr}_{\text{active}} \gets \mathit{ptr}_{\text{clear}}$
        \State $\mathit{ptr}_{\text{clear}} \gets (\mathit{ptr}_{\text{clear}} - 1) \bmod (N_{\text{EBBI}} + 1)$
        \If{$\mathit{ptr}_{\text{clear}} = 0$}  
            \State $ptr_{\text{clear}} \gets N_{\text{EBBI}} + 1$
        \EndIf
        \State $B_{\mathit{ptr}_{\text{clear}}}^+ \gets \mathbf{0}, B_{\mathit{ptr}_{\text{clear}}}^- \gets \mathbf{0}$ \Comment{Clear oldest pair}
        \State $\mathit{event\_count} \gets 0$
        \State $\mathit{t\_start} \gets t_i$
        \EndIf
    \EndProcedure
    
    \State  \textbf{Invariant:} $|\mathcal{S}| = N_{\text{EBBI}} +1$  \Comment{Cardinality constraint}
\end{algorithmic}    
\end{algorithm}

\subsubsection{Patch Extraction From EBBI Stack}

Following the generation and management of the EBBI stack $\mathcal{S}$ (Algorithm \ref{robin}), the next step involves extracting localized spatiotemporal features centered on each incoming event. This process transforms the raw event stream into a fixed-dimensional input vector suitable for the SNN.

For each incoming event $e_i = (x_i, y_i, t_i, p_i)$, an $n \times n$ local patch (where $n$ is odd) centered at $(x_i, y_i)$ is extracted from each of the $N_\emph{EBBI}$ EBBI pairs $(B_k^+, B_k^-)$ in the stack. The cleared pair, which serves as a buffer for future reuse, is excluded from feature extraction. This operation yields two binary matrices $P_k^+, P_k^- \in \{0,1\}^{n \times n}$ for the positive and negative polarity images at each stack index $k$. Patches extending beyond image boundaries are zero-padded.

Each patch matrix is flattened into a vector (row-major order), and the two polarity vectors are concatenated to form a combined binary feature vector $\mathbf{x}_k \in \{0,1\}^{2n^2}$ for each index $k$. Repeating this for all $k = 1, \cdots, N_{\text{EBBI}}$ produces the complete input sequence $\mathbf{X} = [\mathbf{x}_1, \cdots, \mathbf{x}_{N_{\text{EBBI}}}]$. The detailed procedure, including the zero-padding and vectorization, is provided in Algorithm \ref{Patch}.

\begin{algorithm}
    \caption{Patch Extraction From EBBI Stack}
    \label{Patch}
   \begin{algorithmic} [1] 
    \Function{ExtractPatch}{$B, x_c, y_c, n$} 
    \State \textbf{Input:} Binary image $B$ (size $H \times W$), center $(x_c, y_c)$, size $n$ (odd)
    \State \textbf{Output:} $n \times n$ patch matrix $P$
    
    \State $P \gets \text{zeros}(n, n)$ \Comment{Zero-padding}
    \State $r \gets \lfloor n/2 \rfloor$ \Comment{Offset}
    
    \For{$\delta_x \gets -r$ \textbf{to} $r$}
        \For{$\delta_y \gets -r$ \textbf{to} $r$}
            \State $x \gets x_c + \delta_x, \quad y \gets y_c + \delta_y$
            \If{$0 \leq x < W$ \textbf{and} $0 \leq y < H$}  
                \State $P[\delta_x + r, \delta_y + r] \gets B[x, y]$ 
            \EndIf
        \EndFor
    \EndFor
    \State \Return $P$
\EndFunction

\Procedure{ExtractAndVectorize}{$e_i, \mathcal{S}$}
    \State \textbf{Input:} Event $e_i = (x_i, y_i, t_i, p_i)$, EBBI stack $\mathcal{S}$
    \State \textbf{Output:} Feature sequence $\mathbf{X} = [\mathbf{x}_1, \cdots, \mathbf{x}_{N_{\text{EBBI}}}]$
    
    \State $\mathbf{X} \gets [\,]$ \Comment{Initialization: empty list}
    
    \For{$k \gets 1$ \textbf{to} $N_{\text{EBBI}}$}  
        \State $(B_k^+, B_k^-) \gets \mathcal{S}[k]$  
        
        \State \textbf{Step 1: Extract patches}
        \State $P_k^+ \gets \text{ExtractPatch}(B_k^+, x_i, y_i, n)$
        \State $P_k^- \gets \text{ExtractPatch}(B_k^-, x_i, y_i, n)$
        
        \State \textbf{Step 2: Flatten to $n^2$ vectors (row-major} 
        \State $\mathbf{v}_k^+ \gets \text{Flatten}(P_k^+), \quad \mathbf{v}_k^- \gets \text{Flatten}(P_k^-)$
        \State \textbf{Step 3: Concatenate polarity vectors}
        \State $\mathbf{x}_k \gets [\mathbf{v}_k^+; \mathbf{v}_k^-]$  
        
        \State Append $\mathbf{x}_k \text{ to } \mathbf{X}$
    \EndFor
    
    \State \Return $\mathbf{X}$
\EndProcedure
\end{algorithmic}
\end{algorithm}

\subsubsection{Fully Connected Spiking Neural Network (FCSNN) Processing}
\label{sec:fcsnn-processing}

The extracted feature sequence $\mathbf{X} = [\mathbf{x}_1, \cdots, \mathbf{x}_{N_{\text{EBBI}}}]$, where $\mathbf{x}_k \in \{0,1\}^{2n^2}$, is processed by the SNN to produce a classification score, as described in the following section. 

\textbf{Network Architecture.} The proposed FCSNN consists of $L$ layers. The input to the first layer is a $2n^2$-dimensional binary feature vector. This is followed by hidden layers of LIF neurons (where $N_{hidden}^l$ denotes the dimension of the $l$-th layer) that integrate inputs over time to generate spikes. The final layer consists of a single non-LIF neuron that produces a scalar classification score from the final spike pattern, resulting in $L-1$ LIF layers and one linear output layer. 

Eqs. \ref{eq:snn_1} - \ref{eq:snn_3} detail this process, where $x_i(t_s)$ is the $i$-th element of the input vector $X(t_s)$ at timestep $t_s$. Eq. \ref{eq:snn_1} describes the integration and leak of the membrane potential $V_{hidden,j}^l$ for the $j$-th neuron in the $l$-th layer, including a hard reset mechanism. Eq. \ref{eq:snn_2} describes the spike generation occurring when the membrane potential exceeds the threshold $V_{th}$. Here, $x_i^{l-1}(t_s)$ denotes the $i$-th component of the input EBBI vector for $l=1$, or the output spikes from the preceding layer for $l>1$ (i.e. $x_i^{l-1}(t_s)=s_{hidden,i}^{l-1}(t_s)$). Finally, Eq. \ref{eq:snn_3} represents the linear readout operation of the output layer, where $u(\cdot)$ denotes the Heaviside function. The total number of required time steps is $N_{EBBI}+L-2$ to ensure that the final EBBI input propagates through to the output.

\begin{align}
\mathbf{V}_{hidden,j}^{l}(t_s) &= \beta \cdot \mathbf{V}_{hidden,j}^{l}(t_s-1) \cdot \left (1 - s_{hidden,j}^{l}(t_s-1) \right )\notag\\
&+ \sum_{i=1}^{N_{hidden}^{l-1}} w_{ij}^{l}x_i^{l-1}(t_s) \label{eq:snn_1}\\
s_{hidden,j}^{l}(t_s) &= 1 \text{ when } \mathbf{V}_{hidden,j}^{l}(t_s) \geq V_{th} \label{eq:snn_2}\\
y^{last}(t_s) &= u \left (\sum_{j=1}^{N_{hidden}^{last-1}} w_j^{last} s_{hidden,j}^{last-1}(t_s) \right)\label{eq:snn_3}
\end{align}

\subsection{Design Space Exploration}
\label{sec:dse}
The primary design objective is to achieve filtering performance comparable to hardware-friendly alternatives while minimizing implementation complexity. The SNNF performance depends on the following hyperparameters:
\begin{enumerate}
\item EBBI Configuration: The number of EBBI pairs ($N_{\text{EBBI}}$) and the generation trigger, which is either a fixed time interval ($T_e$) or  a fixed event count ($N_e$).
\item SNN Architecture: Network depth, the number of neurons in the hidden layer ($N_{\text{hidden}}$), and the bit-widths for quantized weights and membrane potentials.
\end{enumerate}

The SNNF operates on a local patch of size $n \times n$ ($n = 5$), resulting in two $5 \times 5$ input patches (one per polarity) for each event. Following prior work \cite{Rios-Navarro2023}, we use the DND21 driving dataset due to its challenging nature. The dataset is described in detail in Section \ref{sec:dataset}. Events were split 80\%/20\% for training and testing. For completeness, the results on the hotel-bar dataset are provided in the supplementary material. Following \cite{Guo2022}, \cite{gopalakrishnan2024hast}, the F-1 curve is generated by sweeping the threshold for each parameter choice to obtain a Receiver Operating Characteristic (ROC) curve. The Area Under the Curve (AUC) is used as the primary metric to determine the optimal configuration.

\subsubsection{Selection of EBBI count $N_{\text{EBBI}}$}
\label{subsec:ebbi_count}

\begin{figure}[htb]
\centering
\captionsetup[subfigure]{font=scriptsize, labelfont=scriptsize}
  \begin{subfigure}{0.4\textwidth}
    \centering
    \includegraphics[width = \linewidth]{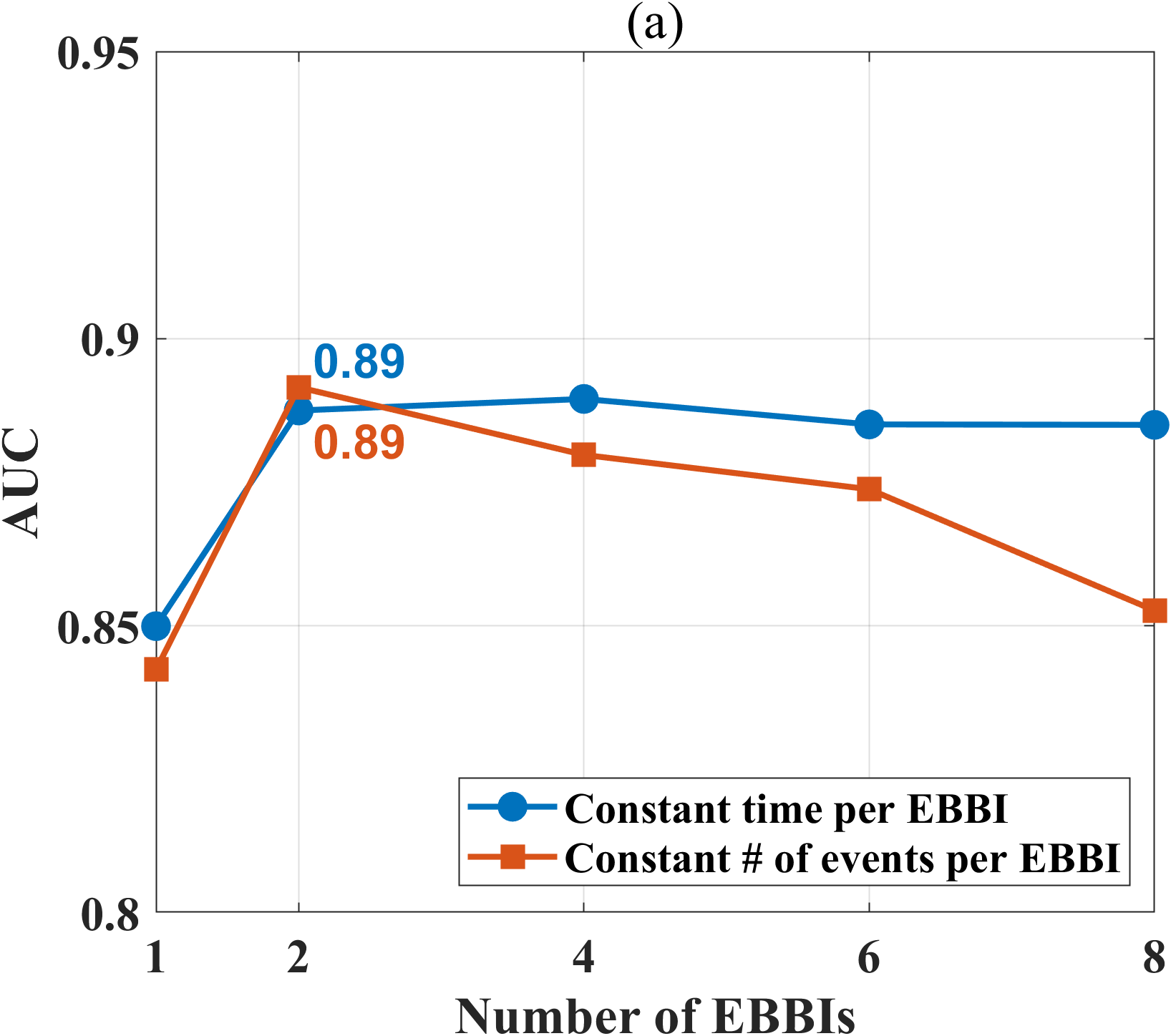}
    \label{fig:compA}
  \end{subfigure}
  
  \vspace*{-0.3cm}
  
  \begin{subfigure}{0.4\textwidth}
    \centering
    \includegraphics[width = \linewidth]{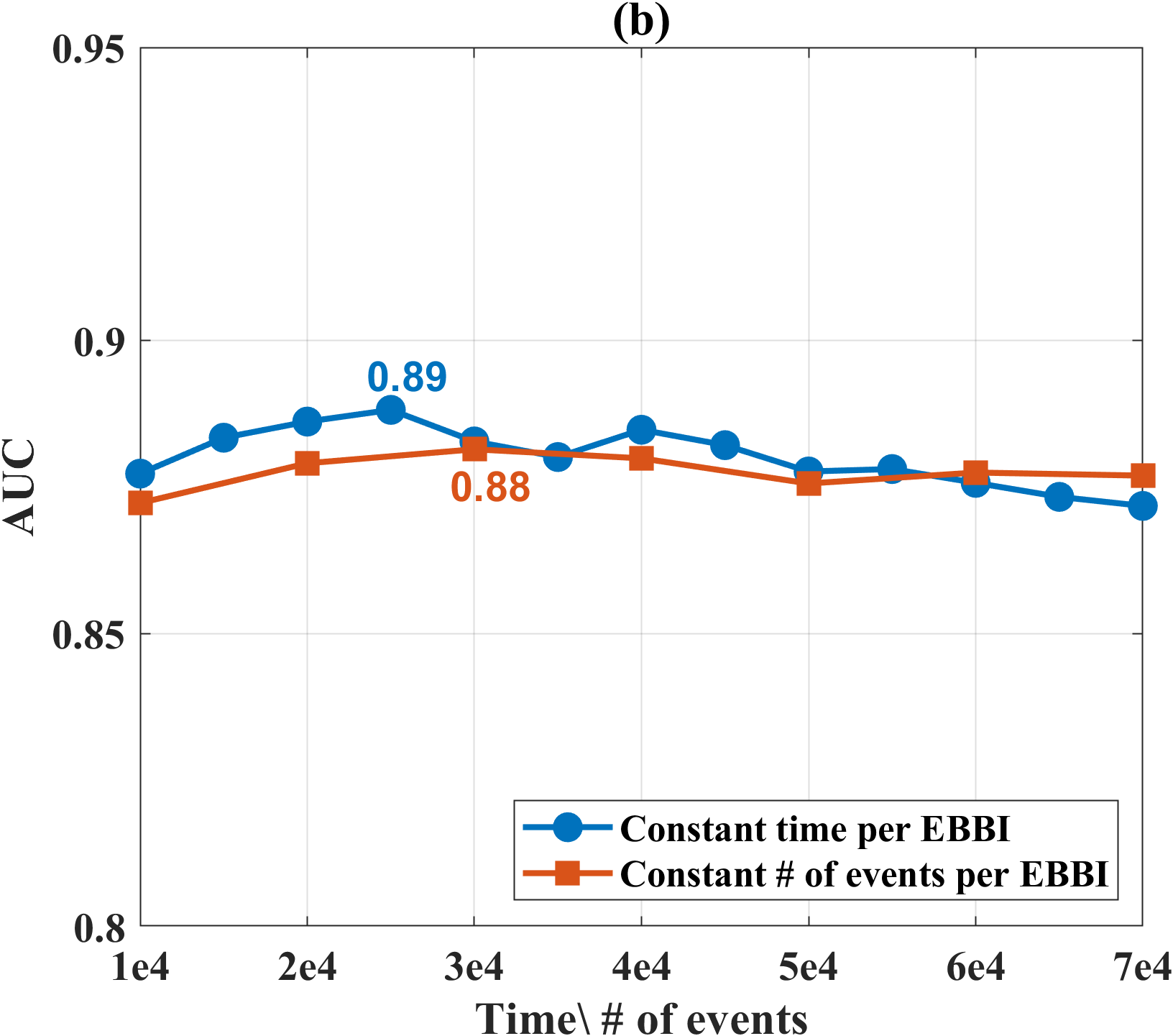}
    \label{fig:compB}
  \end{subfigure}
\caption{Performance of the 1-Layer FCSNN filter: (a) AUC vs. number of EBBI pairs ($N_{\text{EBBI}})$, (b) AUC vs. the EBBI creation hyperparameters: time interval $T_e$ and event count $N_e$}
\label{fig:1L_performance}
\end{figure}

EBBIs provide a condensed spatiotemporal representation of the neighborhood around incoming events. The count $N_{\text{EBBI}}$ critically influences both filtering performance and hardware resource utilization, specifically memory footprint and throughput.
We performed multiple parameter sweeps on the driving dataset to evaluate AUC as a function of $N_{\text{EBBI}}$. During this sweep, the architecture was fixed to a single hidden layer ($L=2$), matching the optimal MLP architecture in \cite{Guo2022}. This shallow structure reduces processing latency, which is crucial for real-time event camera denoising. Fig. \ref{fig:1L_performance}(a) shows the AUC versus the number of EBBI pairs for the DND21 dataset with shot noise. Across both generation methods, optimal performance is achieved with $N_{\text{EBBI}} = 2$ (4 EBBIs total).

\subsubsection{Selection of EBBI creation method}
Beyond the count, we evaluated whether the EBBI creation should be triggered by a fixed time interval $T_e$ or a fixed event count $N_e$. As shown in Fig. \ref{fig:1L_performance}(a), both methods achieve a maximum AUC of 0.89. However, the constant-time method yields higher AUC more consistently across the parameter range. Fig. \ref{fig:1L_performance}(b) shows a detailed hyperparameter sweep on the driving dataset. It confirms that constant-time generation outperforms constant-event generation, with a peak AUC of 0.89 at $T_e = 25$ ms compared to 0.88 at $N_e = 30,000$. Consequently, we adopt the constant-time EBBI creation method with $T_e = 25$ ms.

\subsubsection{Selection of SNN Architecture}
\label{subsec:snn_architecture}

\begin{figure}[htb]
\centering
\captionsetup[subfigure]{font=scriptsize, labelfont=scriptsize}
  \begin{subfigure}{0.41\textwidth}
    \centering
    \includegraphics[width = \linewidth]{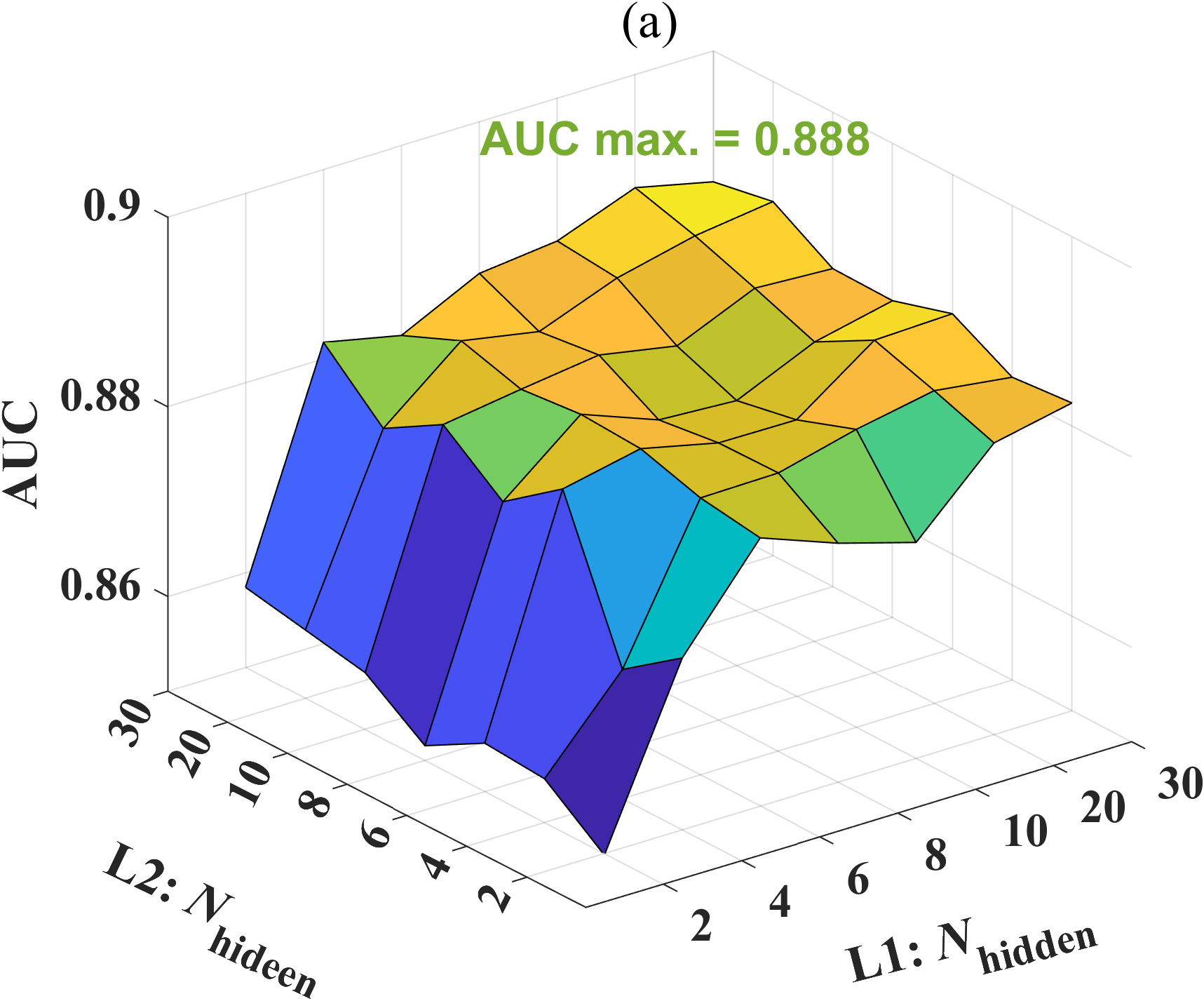}
    \label{fig:compB}
  \end{subfigure}
  
  \vspace*{-0.3cm}
  
  \begin{subfigure}{0.4\textwidth}
    \centering
    \includegraphics[width = \linewidth]{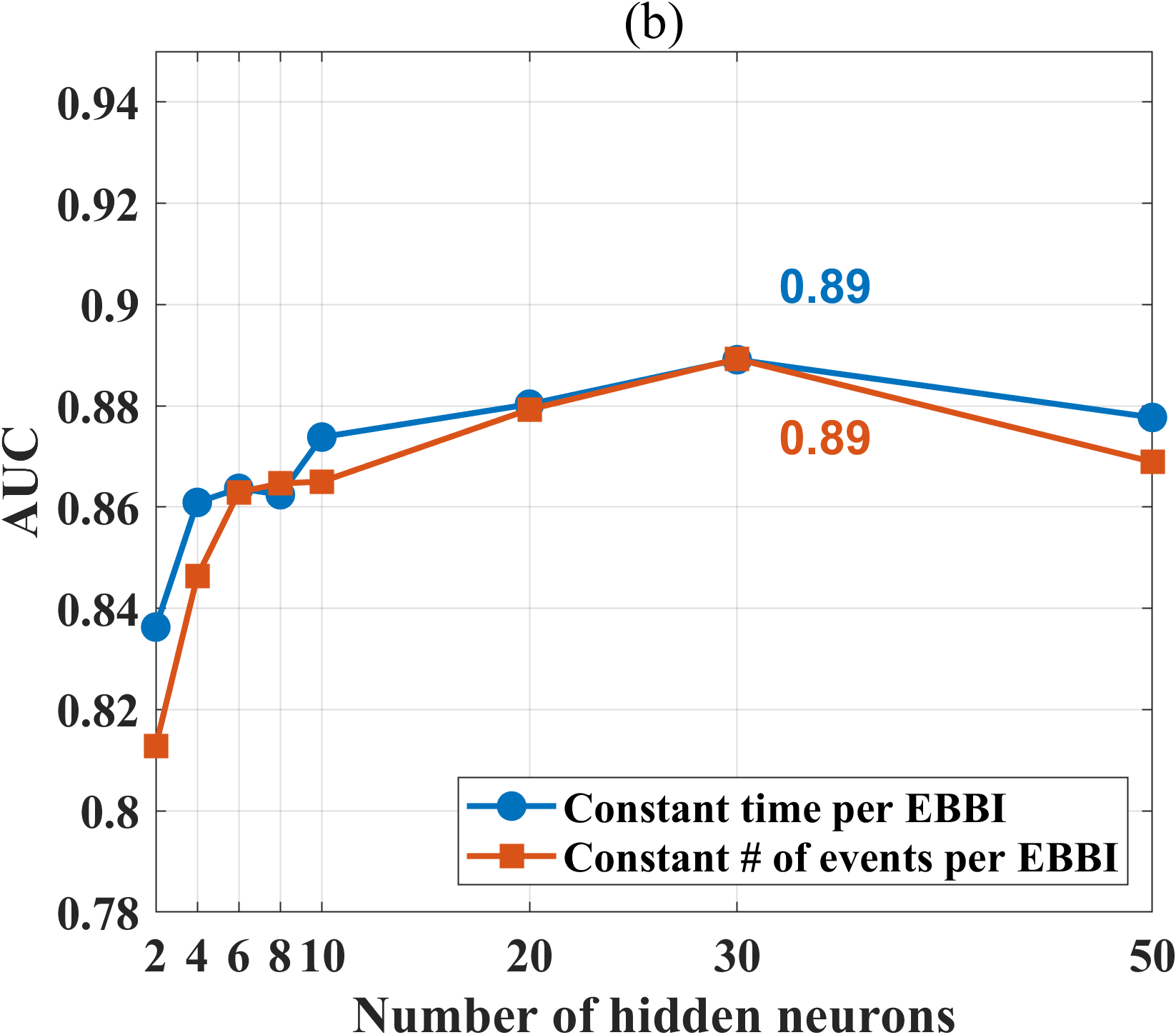}
    \label{fig:compA}
  \end{subfigure}
\caption{(a) Performance of the 2-Layer FCSNN filter: AUC vs. number of neurons in each layer. (b) Performance of the 1-Layer FCSNN filter: AUC vs. number of hidden neurons ($N_{\text{hidden}}$). }
\label{fig:num_neurons}
\end{figure}

To validate our architectural assumption, we evaluated three candidates: a 1-layer FCSNN ($L=2$), a 2-layer FCSNN ($L=3$), and a Convolutional Spiking Neural Network (CSNN). The 1-layer FCSNN performed on par with  deeper networks (Fig. \ref{fig:num_neurons}(a)) and was hence selected for its optimal balance of performance and hardware efficiency. Convolutional architectures did not yield significant filtering performance gains and were excluded due to their higher hardware complexity. The number of hidden neurons was determined via a sweeping shown in Fig \ref{fig:num_neurons}(b); a 30-neuron configuration ($N_\text{hidden}^1=30$) was chosen as further increases yielded diminishing returns. Finally, weight and membrane potential quantization were set to $8$ and $12$ bits, respectively, based on the optimal AUC.

\section{Hardware Architecture}
\label{sec:hardware_arch}

The overall architecture of the SNNF is illustrated in Fig. \ref{fig:arc}(b). The system receives an asynchronous event stream from a DVS to construct and update $N_{\text{EBBI}}$ pairs of EBBIs in memory, where each EBBI frame matches the DVS sensor resolution of $W \times H$ pixels. For every event, an $n \times n$ pixel patch is extracted from the $N_{\text{EBBI}}$ EBBI pairs surrounding the event location $(x, y)$. This patch data forms the input to a single-hidden-layer FCSNN with $N_{\text{hidden}}$ neurons. Key hardware implementation parameters are summarized in Table \ref{tab:specifications}, with details provided in the remainder of this section.

\begin{table}[!htbp]
\centering
\caption{Hardware SNNF specifications.}
\label{tab:specifications}
\begin{tabularx}{\linewidth}{>{\centering\arraybackslash}X|>{\centering\arraybackslash}X|>{\centering\arraybackslash}X}
\toprule
 Parameter & Description & Value \\ 
 \toprule
 $W \times H$ & DVS width×height & $346 \times 260$ \\ 
 $n \times n$ & input patch & $5 \times 5$ \\
 $N_\text{EBBI}$ & \# EBBI pairs & 2 \\
 $N_\text{hidden}$ & \# hidden neurons & 30 \\
 $T_\text{e}$ & fixed time & 25 ms \\
 $N_\text{mem}$ & \# memory banks & 5 \\
 $W_\text{word}$ & word width in banks & 8 bits \\ 
\bottomrule
\end{tabularx}
\end{table}

\subsubsection{Memory bank architecture}
\label{sec:mem}
\begin{figure} [!hthp]
\centering
\includegraphics[ width= 0.48 \textwidth]{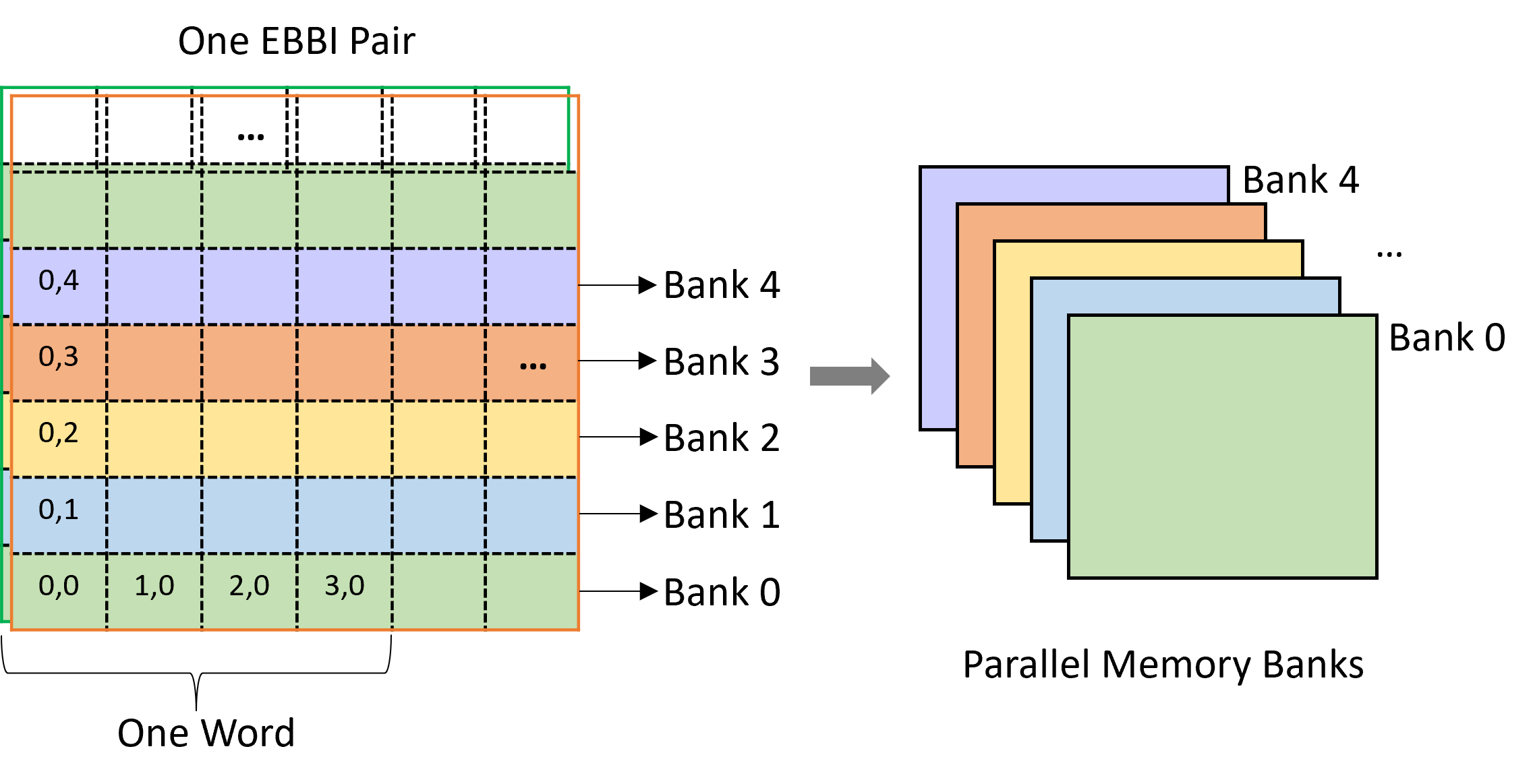}
\caption{Parallel memory bank architecture for storing EBBI pairs. The image is striped across multiple banks (Bank 0–4) to enable parallel access to local patches}
\label{fig:mem}
\end{figure}

Patch extraction, as outlined in Algorithm \ref{Patch}, requires reading $n^2$ pixels. A standard single-port memory would require at least $n^2$ clock cycles to complete this operation, creating a throughput bottleneck. To enable high-throughput operation, we employ a parallel memory architecture. As shown in Fig. \ref{fig:mem}, the pixel data of a $W \times H$ EBBI is distributed across $N_\text{mem}$ memory banks (e.g., Bank 0 to Bank 4). Image rows are divided into $N_\text{mem}$ contiguous blocks, with each bank storing one block of approximately $\lceil H/N_\text{mem} \rceil$ rows. This striping pattern allows an $n \times n$ patch, which spans at most $n$ consecutive image rows, to be fetched by accessing the corresponding rows across $n$ banks simultaneously. To minimize hardware complexity while enabling parallel retrieval, we set $N_\text{mem} = n$. Each addressable word in a bank stores $W_\text{word}$ bits of pixel data. This bank-level parallelism could enable the complete input patch to be retrieved in a single cycle, drastically improving system throughput while reducing access latency.

The choice of $W_\text{word}$ could also impact performance, hardware efficiency and implementation complexity. While bank-level parallelism allows $n$ rows to be accessed simultaneously, throughout is also dictated by the memory word width $W_\text{word}$ and data alignment. Since each pixel is represented by a single bit, the $n$ consecutive pixels in a row may straddle word boundaries. In the worst-case, these $n$ pixels could span $\lceil n/{W_\text{word}} \rceil$ adjacent words, requiring sequential reads from the same bank, which increases the access latency.

 We evaluated three candidate widths for our $n=5$ design: 
 \begin{itemize}
 \item \textbf{2-bit:} Maximizes memory efficiency but requires up to $\lceil 5/2 \rceil = 3$ reads per row in the worst case, degrading throughput. 
 \item \textbf{8-bit:} Allows a 5-pixel slice to fit within one read in the best case, but the worst case still requires 2 reads (16 bits fetched, 5 used), resulting in 68.75\% bandwidth waste. 
 \item \textbf{4-bit:} Represents the optimal trade-off. The worst case requires 2 reads to cover 5 pixels (8 bits total), with only 3 (37.5\%) unused bits. This configuration minimizes the data transfer redundancy.
 \end{itemize}

Therefore, we select $W_\text{word} = 4$ bits. With $N_\text{mem}=5$ banks, the system reads $5 \times 4 = 20$ bits per cycle. In the worst-case alignment, fetching a complete $5 \times 5$ patch requires 2 cycles (40 bits in total), from which the required 25 pixels are selected. Compared to a serial access pattern requiring $n^2 = 25$ cycles, this parallel architecture ($N_{\text{mem}}=5$ and $W_{\text{word}}=4$) reduces the worst-case latency to 2 cycles, achieving a $12.5\times$ speedup critical for real-time event processing.

\subsubsection{System dataflow}
\label{sec:dataflow}
\begin{figure*} [!hthp]
\centering
\includegraphics[ width= 0.95 \linewidth]{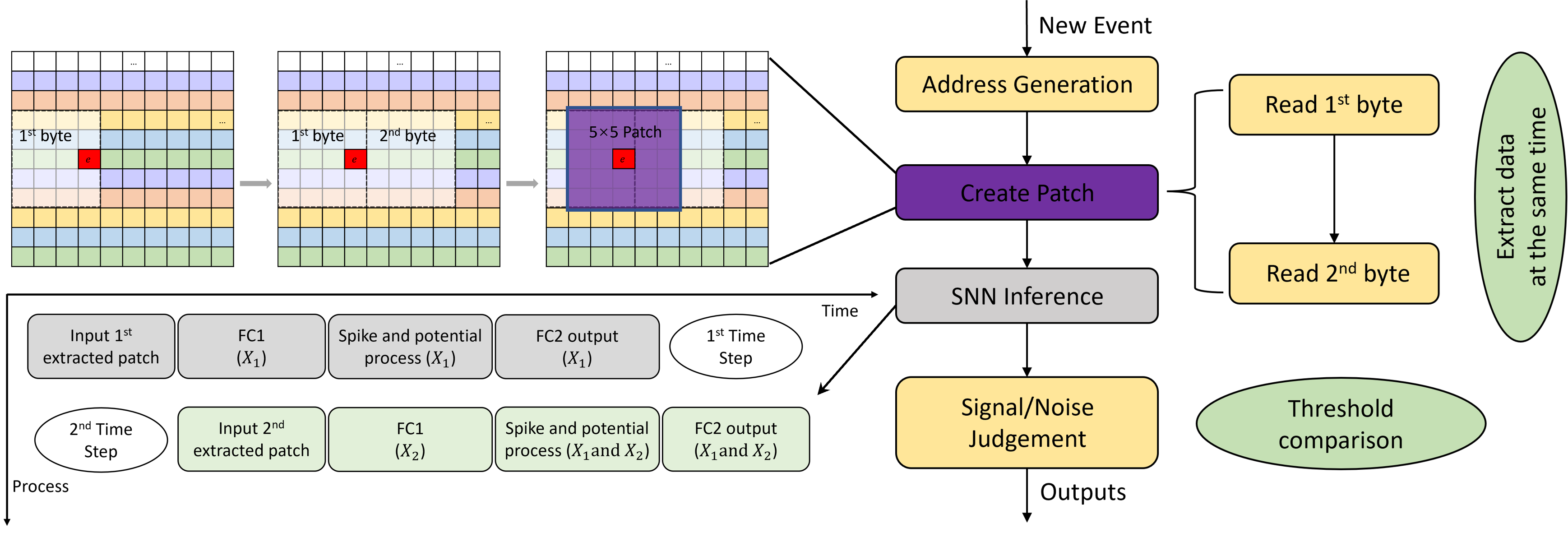}
\caption{System-level dataflow and timing diagram of the SNNF. The upper-left section depicts patch extraction, the lower-left section illustrates the five-stage FCSNN processing pipeline, and the right side shows the complete end-to-end processing flow.}
\label{fig:flow}
\end{figure*}

The complete system dataflow of the SNNF is illustrated in Fig. \ref{fig:flow}. The architecture follows a serial design that processes incoming events through four main stages, orchestrated by a finite-state machine (FSM) controller: (1) address generation for input events, (2) patch extraction from the EBBI stack, (3) feature processing through the FCSNN, and (4) classification. The core components of the filter are the FSM controller, the parallel memory banks, and the FCSNN.

\textbf{Stage 1: Address Generation.} The system receives asynchronous events from the DVS. For each incoming event $e_i = (x_i, y_i, t_i, p_i)$, the hardware generates multiple memory addresses in parallel, including: (a) the write address to update the corresponding pixel in the active EBBI pair based on the event polarity, and (b) the read addresses for extracting the surrounding $5 \times 5$ patch from the relevant memory banks. Addresses for neighboring pixels required during the two-cycle patch extraction are pre-computed to maximize throughput.

\textbf{Stage 2: Patch Extraction.} This stage requires two clock cycles. In the first cycle, the system retrieves 20 bits (the first set of 4-bit words) from the $N_{\text{mem}}=5$ parallel memory banks. In the second cycle, it reads the adjacent words to retrieve another 20 bits. From these 40 bits, the 25 relevant pixels (5 rows $\times$ 5 columns) of the patch are selected and the rest are discarded. The extracted patches from both polarities are flattened and concatenated into a 50-dimensional binary feature vector, producing a temporal sequence of $N_{\text{EBBI}}$ feature vectors.

\textbf{Stage 3: FCSNN Processing.} The sequence of feature vectors is processed by a five-stage FCSNN pipeline (lower-left of Fig.~\ref{fig:flow}): (1) input buffering, (2) FC1 layer computation (50-dimensional vectors to $N_{\text{hidden}}$-dimensional currents), (3) LIF neuron spike generation and membrane potential update, (4) FC2 layer computation (scalar score), and (5) Output buffering.

The pipeline enables concurrent processing of successive inputs: while the $i$-th input undergoes LIF processing (integrates currents from previous timesteps), the $(i+1)$-th input is processed by the FC1 stage. Governed by Eq. (\ref{eq:snn_1}), the LIF neurons accumulate currents across $N_{\text{EBBI}}$ timesteps. Consequently, processing a full sequence requires $N_{\text{EBBI}} + 3$ clock cycles. As shown in Fig. \ref{fig:flow}, our system uses $N_{\text{EBBI}}=2$.

\textbf{Stage 4: Classification and Output.} The FC2 layer computes a scalar value $y$ from the final spike pattern, which is compared against a predefined threshold $V_{th}$ (right side of Fig.~\ref{fig:flow}). If $y \geq V_{th}$, the event is classified as \textit{signal}; otherwise, it is \textit{noise}. The binary decision is output alongside the event timestamp and coordinates. $V_{th}$ is tuned during training to balance detection sensitivity and false positive rate.

\textbf{End-to-End Latency.} The total latency from event arrival to classification is deterministic. It comprises the sequential execution of Stage 1 (1 cycle), Stage 2 (2 cycles), Stage 3 FCSNN pipeline ($N_{\text{EBBI}}+3=5$ cycles for $N_{\text{EBBI}}=2$), and Stage 4 comparison (1 cycle), totaling 9 clock cycles. Once the pipeline is filled, the system achieves a throughput of one classification per cycle. This deeply pipelined dataflow, combined with parallel memory access and optimized FCSNN processing, enables the high-throughput, low-latency processing essential for real-time neuromorphic vision applications.

\section{Results}
\label{sec:results}

\subsection{Methods}
\subsubsection{Metrics}
A binary classifier labels each incoming event as either signal or noise; accuracy is determined by comparing its output against a \textit{Ground Truth} (GT). A \textit{False Positive} (FP) occurs when an event identified as noise in the GT is classified as signal by the filter, whereas a \textit{False Negative} (FN) occurs when a signal event is classified as noise. To evaluate the filter performance, we utilize the \textit{False Positive Rate} (FPR) and \textit{True Positive Rate} (TPR). FPR is defined as $FP/N_N$ where $N_N := FP+TN $ is the total number of negative samples (noise events). TPR is defined as $TP/N_P$ where $N_P := TP+FN$ is the total number of positive samples (signal events). The ROC curve combines these metrics to assess overall system performance. The AUC, ranging from 0 to 1, quantifies this performance. A higher AUC value indicates better performance; an AUC of 1 represents a perfect classifier. 

\subsubsection{Performance evaluation methodology}
\label{Performance Evaluation Methodology}

We used a combination of Python and MATLAB simulation environments to evaluate the BA de-noising capability of the proposed filter against other hardware-friendly benchmarks. A Python model of the proposed filter was developed based on the architecture in Section \ref{sec:hardware_arch}, using the \textit{snnTorch} framework \cite{Eshraghian2023} to model, train, and test the FCSNN. Section \ref{sec:dse} describes the methodology used to optimize the SNNF configuration to maximize filtering performance while minimizing hardware resource utilization. 

MATLAB models for the comparison filters were implemented based on their descriptions in \cite{Liu2015}, \cite{Guo2021}, and
\cite{Khodamoradi2021}. For the HashHeat filter\cite{Guo2021}, we applied parameter values estimated to maximize filter performance (see Section \ref{sec:hashheat}). We compared the SNNF filter against these filters using two distinct datasets. Event streams were processed through the MATLAB/Python models, and performance was quantified by plotting ROC curves and calculating the AUC. 

\subsubsection{Dataset}
\label{sec:dataset}
The \textit{driving} dataset from DND21 \cite{Guo2022} was generated using the \textit{Video to Events} (v2e) tool \cite{Hu2021}, which converted a simulated video of resolution $346$ $\times$ $260$ pixels (scenes from the dashboard camera of a car driven through a city) into approximately $3.9$M events over $6$ seconds. This dataset represents a typical moving-camera DVS application. 
To simulate realistic conditions, noise-free events from the \textit{driving} dataset were mixed with recorded \textit{leak noise} and \textit{shot noise} noise from DND21. These noises are recorded using a Davis346 camera under high and low intensity illumination, respectively. In this setup, GT defines all original (noise-free) events from the \textit{driving} dataset as signals and all added recorded events as noise. An event is classified as TP/TN/FP/FN based on how it is labelled by and defined in the filter and GT, respectively, from which the TPR and FPR values can be calculated. This is similar to the method used in \cite{gopalakrishnan2024hast}.

\subsection{Algorithm results}
\label{sec:performance_analysis}
The methodology in Section \ref{Performance Evaluation Methodology} was used to evaluate the proposed SNNF against three hardware-friendly benchmarks in Section \ref{BA_filters}: MLPF, Guo-STCF, and BAF. These filters were chosen for their high performance on the \textit{Driving} dataset (results for the hotel-bar dataset are reported in the Supplement).
\subsubsection{Filter Accuracy Analysis}
\begin{figure}[htb]
\centering
\captionsetup[subfigure]{font=scriptsize, labelfont=scriptsize}
\includegraphics[width=0.45 \textwidth]{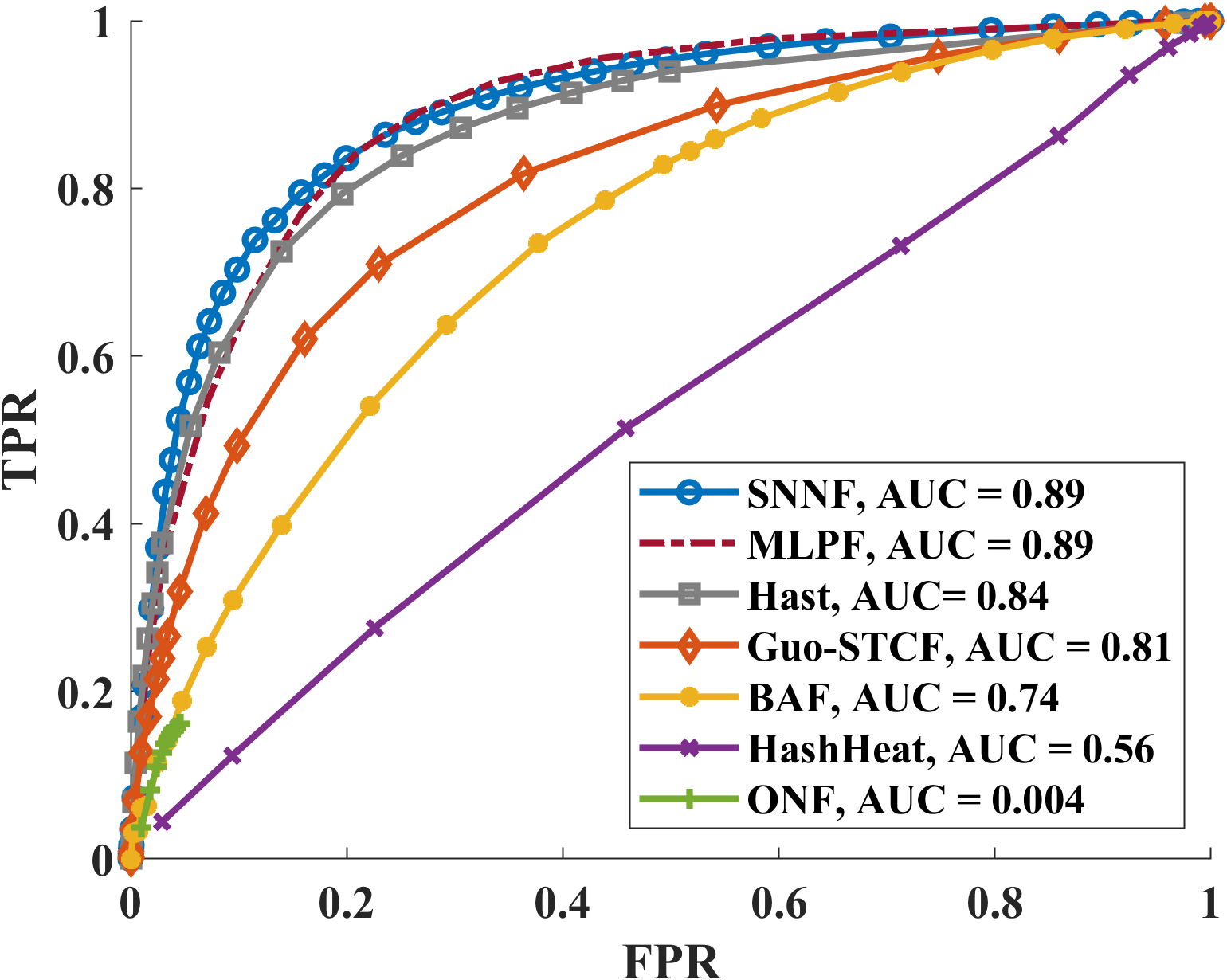}
\caption{ROC curve and AUC of \FNAME{} versus other filters for \textit{driving} data mixed with \textit{shot noise}. The proposed SNNF demonstrates the highest filtering accuracy.}
\label{fig:fpr_vs_tpr}
\end{figure}

Following \cite{Guo2021}, the ROC curves for Guo-STCF and BAF were plotted by varying $\tau$ from $100$ $\mu$s to $1$s. For Guo-STCF, an event was declared as a signal if four supporting events were present ($k = 4$). For SNNF and MLPF, ROC curves were plotted by varying the comparison thresholds at the final network stage (see Fig. \ref{fig:arc}(b)). Fig. \ref{fig:fpr_vs_tpr} shows the ROC curves for these filters using \textit{driving} data from DND21 mixed with \textit{shot noise}. The ROC curve of MLPF shown in the figure is simulated using an unquantized 32-bit floating-point model. The 4-bit quantized AUC of MLPF on this dataset is 0.87 \cite{Rios-Navarro2023}. In contrast, the \FNAME{} simulation used a hardware-accurate quantized model with 8-bit weights and 12-bit membrane potentials detailed in Section \ref{sec:hardware_arch}. Despite this quantization, \FNAME{} marginally outperforming MLPF, while Guo-STCF and BAF exhibited 6--7\% and 11--13\% lower AUC values, respectively.

\subsubsection{Filter Resource Utilization Analysis}
\begin{figure*}[htb]
\centering 
\begin{minipage}{1\linewidth}	
\captionsetup[subfigure]{font=scriptsize, labelfont=scriptsize}
\subfloat[] {\includegraphics[width=0.33 \textwidth]{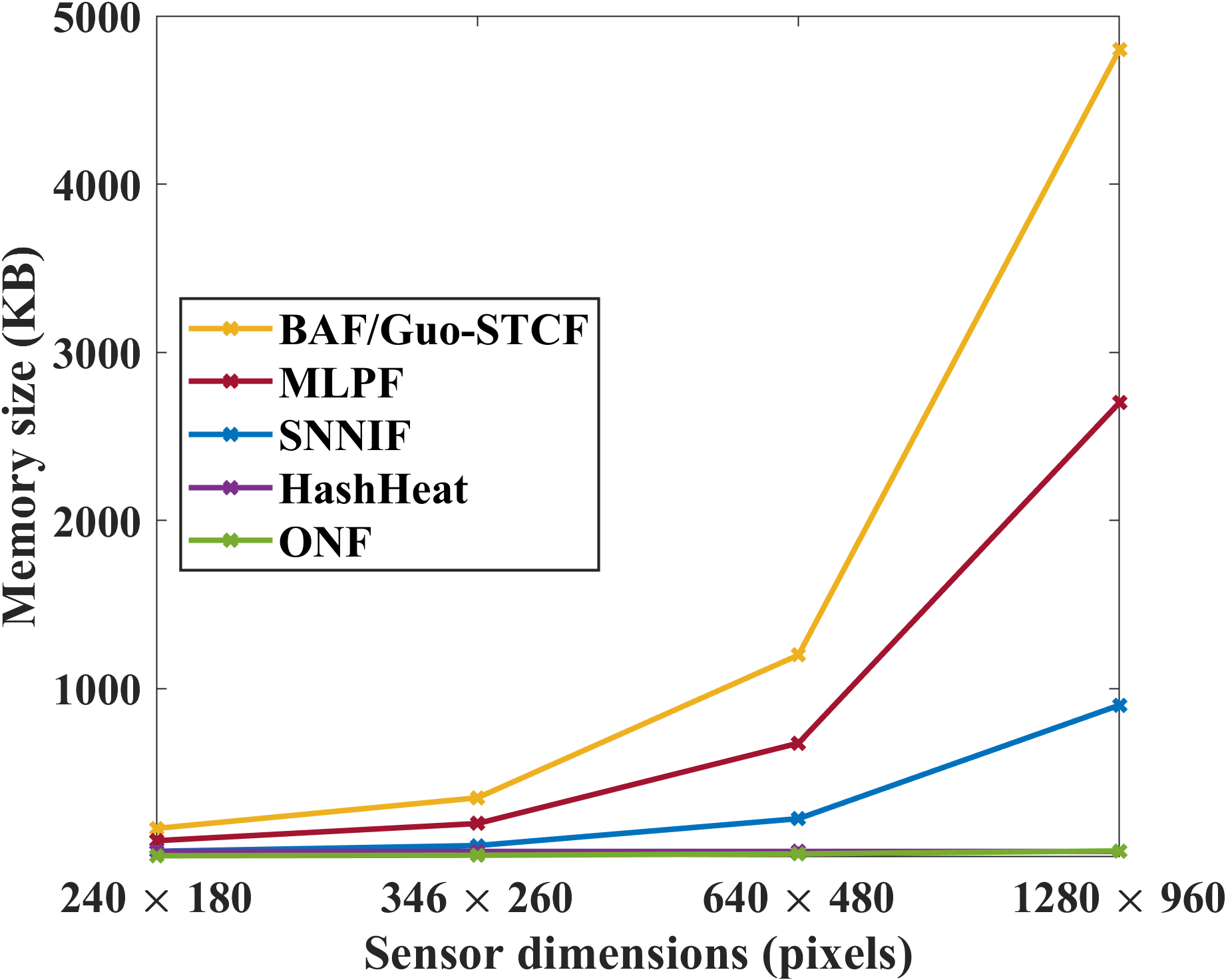}} 
\subfloat[] {\includegraphics[width=0.33\textwidth]{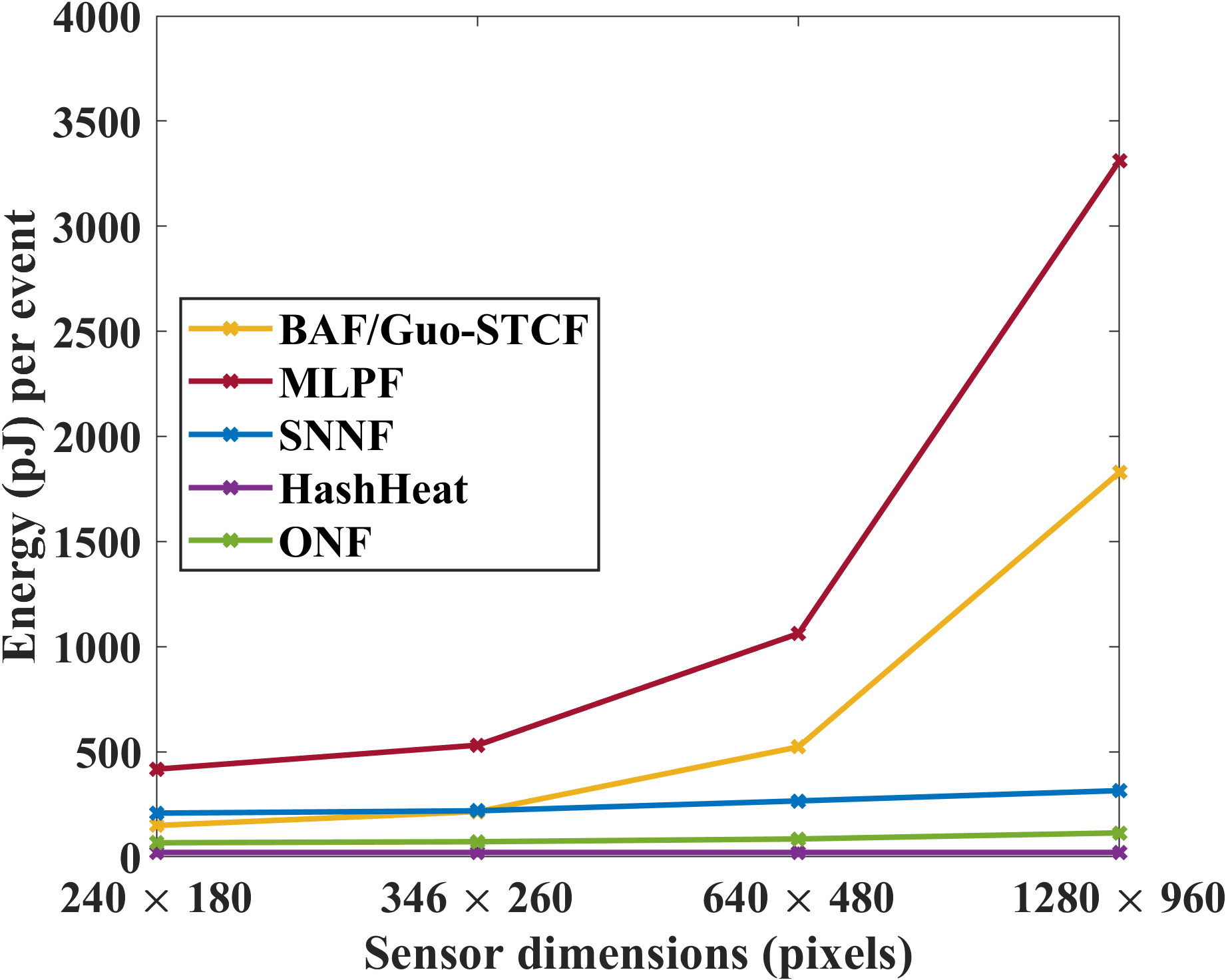}} 
\subfloat[] {\includegraphics[width=0.33 \textwidth]{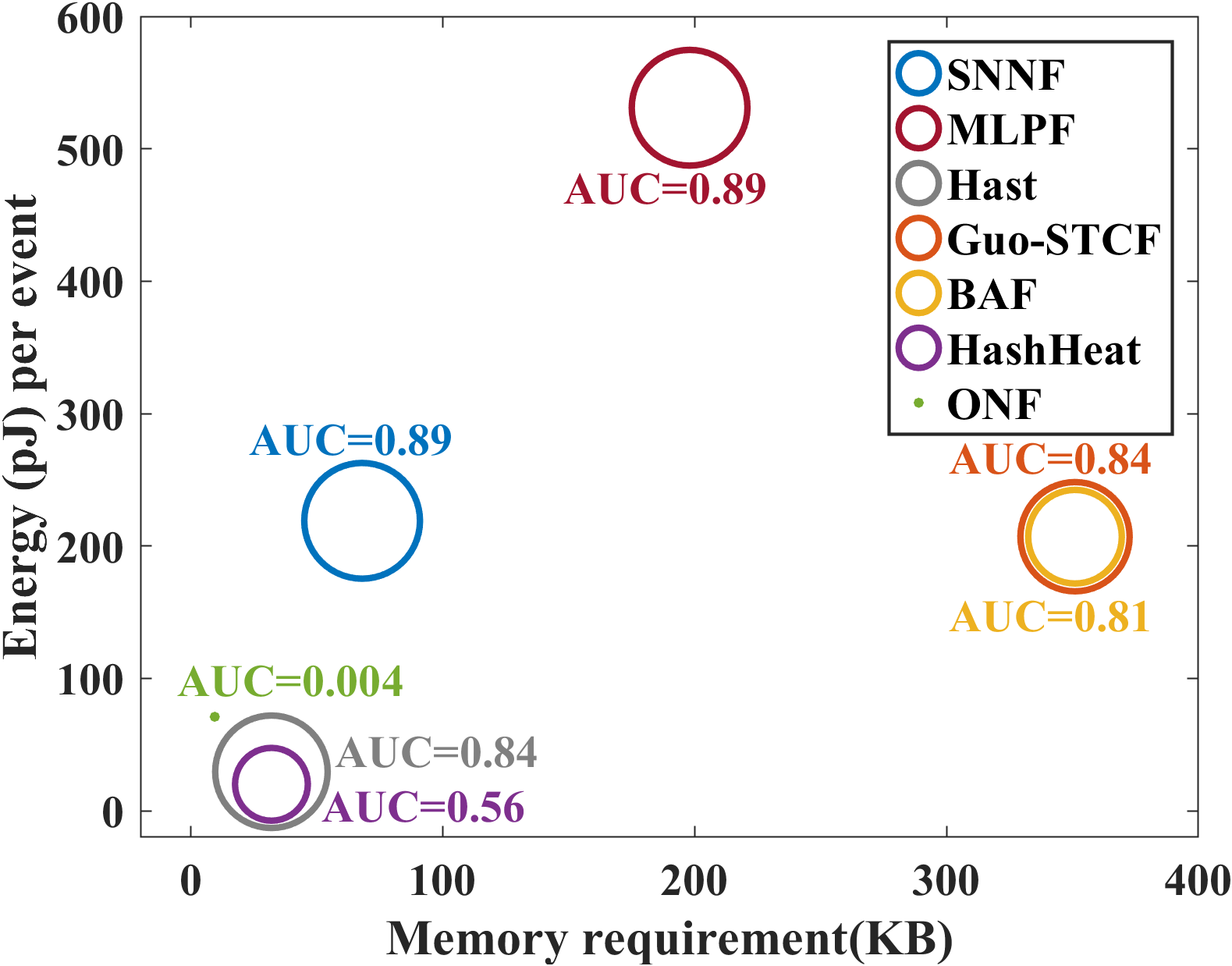}} 
\end{minipage}
\caption{Scaling of memory requirement and estimated energy per event in picojoules (pJ) with increasing sensor resolution for \FNAME{} as compared to BAF/Guo-STCF and MLPF. 
(a) Memory requirement (b) Estimated energy per event in pJ, (c) Memory requirement, energy consumption per event, and AUC, derived from simulations using the \textit{driving} dataset mixed with \textit{shot} noise; sensor resolution = $346$ $\times$ $260$ pixels.}
\label{fig:mem_pj_vs_sensor_dim} 
\end{figure*}

Hardware efficiency was evaluated based on \textit{memory footprint} and \textit{energy per event} \cite{gopalakrishnan2024hast}. Fig. \ref{fig:mem_pj_vs_sensor_dim} shows the comparison of different filters.

Memory scaling varies significantly by filter type, as shown in Fig. \ref{fig:mem_pj_vs_sensor_dim}(a). Guo-STCF and BAF \cite{Delbruck2008} require $32 \times W \times H$ bits, while ONF \cite{Khodamoradi2021} requires $64\times (W+H)$ bits. HashHeat \cite{Guo2020} employs a fixed memory size of $128 \times 16$ bits \cite{Rios-Navarro2023}. Conversely, SNNF's memory scales  primarily with EBBI count and sensor resolution, requiring only about 13$\%$ of the memory used by BAF/Guo-STCF and about 33$\%$ of that used by MLPF at any resolution.

The energy costs depicted in Fig. \ref{fig:mem_pj_vs_sensor_dim}(b) were estimated by summing fundamental operation costs from \cite{Horowitz2014} in the same manner as \cite{gopalakrishnan2024hast}. Although HashHeat and ONF are more energy-efficient, they show poor filtering accuracy. MLPF consumes the highest energy per event due to 48 18-bit memory reads per event. SNNF requires only 10 reads per event, and its 4-bit memory access significantly reduces energy overhead. Most energy was consumed in FCSNN classification, with minimal increase at higher resolutions due to low memory overhead. 

To provide a holistic comparison, Fig. \ref{fig:mem_pj_vs_sensor_dim}(c) maps the tradeoff between filtering performance (AUC), energy consumption per event, and memory requirement of the filters in comparison. Each circle represents a filter. The area of each circle denotes the AUC simulated using the \textit{driving} dataset mixed with \textit{shot} noise, for a sensor resolution of 346 $\times$ 260 pixels. The \textit{x} and \textit{y} coordinates of the center of each circle represent the memory requirement in kilobytes (KB) and energy consumption in picojoules (pJ) per event, respectively. Circles closer to the origin indicate higher hardware efficiency (i.e., lower energy and memory consumption), and a larger circle means higher filtering performance. As BAF and Guo-STCF have similar energy and memory requirements, the centers of their circles coincide. While MLPF matches \FNAME{} in accuracy, its energy consumption is significantly higher. ONF and HashHeat are closer to the origin but have substantially smaller areas, indicating inferior filtering performance. These results demonstrate that \FNAME{} provides the best balance of accuracy and resource efficiency for near-sensor applications.

\subsection{Hardware results}
\label{sec:hardware_result}
The proposed SNNF architecture was implemented on FPGAs and post-layout simulated using a standard ASIC design flow.
\subsubsection{FPGA Implementation}
Two FPGA platforms were used for implementation. To ensure a fair comparison with \cite{Rios-Navarro2023}, we targeted the Xilinx Zynq XC7Z100 and Zynq Ultrascale+ ZU3CG. The core filter logic comprising the finite-state machine controller, parallel memory banks, and the FCSNN classifier (described in Section \ref{sec:dataflow}) was implemented in Verilog HDL. The design was synthesized, placed, routed, and simulated using the AMD-Xilinx Vivado toolchain. Quantized weights from the Python/snnTorch training phase were embedded directly into the hardware. Filter parameters remained consistent with the \textit{driving} dataset experiments described in Section \ref{sec:performance_analysis}.


\begin{table}[htbp]
\centering
\begin{threeparttable}
\caption{Performance metrics for FPGA implementation}
\label{tab:FPGA}
\begin{tabular}{llccccc}
\toprule
\textbf{Platform}& \textbf{Filter}&\textbf{LUT}&\textbf{FF}&\textbf{BRAM}&\textbf{Freq.}&\textbf{MER}\\ 
& &\textbf{(\%)} &\textbf{(\%)} & \textbf{(\%)}&\textbf{(MHz)}&\textbf{(Meps)}\\ 
\midrule
\multirow{2}{*}{\textbf{XC7Z100}} & MLPF [1] & 6 & 0.72 & 26 & 100 & 10 \\
& SNNF & 2.54 & 0.38 & 2.98 & 211 & 23.44 \\
\cmidrule(r){1-7}
\multirow{2}{*}{\textbf{ZU3CG}} & MLPF [1] & 34 & 2 & 92 & 236 & 23.6 \\
& SNNF & 9.46 & 1.49 & 10.42 & 261 & 29 \\
\bottomrule
\end{tabular}
\begin{tablenotes}
\item[1] MER: Maximum Event Rate
\item[2] LUT: Look-Up Table utilization
\item[3] FF: Flip-Flop utilization
\item[4] BRAM: Block RAM utilization
\item[5] Meps: Million events per second
\end{tablenotes}
\end{threeparttable}
\end{table}


Table \ref{tab:FPGA} summarizes the resource utilization, maximum clock frequency ($f_\emph{clk}$), and maximum event rate (MER) achieved for the XC7Z100 and ZU3CG implementations of SNNF. For context, results for the MLPF filter from \cite{Rios-Navarro2023} on the same platforms are included. Note that BAF and HashHeat results are omitted as their implementations on these specific devices are unavailable. Memory requirements are reported for a sensor resolution of $346 \times 260$ pixels.

On the XC7Z100, the SNNF achieves an $f_\emph{clk}$ of 211.3 MHz, compared to 100 MHz for MLPF. On the more advanced UltraScale+ ZU3CG, it reaches 261.09 MHz versus 236 MHz for MLPF. As detailed in Section \ref{sec:dataflow}, processing a single event requires a deterministic pipeline of 10 clock cycles: 2 cycles for patch extraction and 8 cycles for buffering and FCSNN inference. Consequently, the throughput is calculated as $f_{clk}$/10, resulting in 23.44 Meps on the XC7Z100 and 29.01 Meps on the ZU3CG--both significantly higher than MLPF's throughputs of 10 Meps and 23.6 Meps, respectively.

Resource utilization also favors the SNNF. Look-Up Table (LUT) usage is less than half that of MLPF on both FPGAs. Flip-Flop (FF) usage is less than 53\% of MLPF's requirement on the XC7Z100. Most notably, SNNF's Block RAM (BRAM) consumption is only 2.98\% on the XC7Z100 and 10.42\% on the ZU3CG, compared to 26\% and 92\%, respectively for MLPF. This reduction (below 12\% of MLPF's footprint) stems from our parallel memory architecture (Section \ref{sec:mem}), which decouples memory size from sensor resolution. Unlike MLPF, which requires one memory bank per sensor column ($\sim881$ KB or 400 BRAMs for a $346 \times 260$ sensor to minimize latency arising from the search operation), SNNF is highly scalable for resource-constrained, near-sensor processing.

In summary, the SNNF demonstrates superior hardware efficiency, as evinced by its markedly lower hardware implementation complexity and a more favorable performance per resource trade-off than the MLPF benchmark on both FPGA platforms.

\subsubsection{ASIC Implementation}
\label{subsec:asic_impl}

To evaluate the filter performance as a custom silicon chip, the design was migrated to a TSMC 65 nm process. Parallel memory banks were implemented using custom single-port SRAM IP cores. The design was synthesized with Cadence and placed-and-routed using Cadence Innovus.

\begin{figure} [htbp]
\centering \includegraphics[width=0.6\columnwidth]{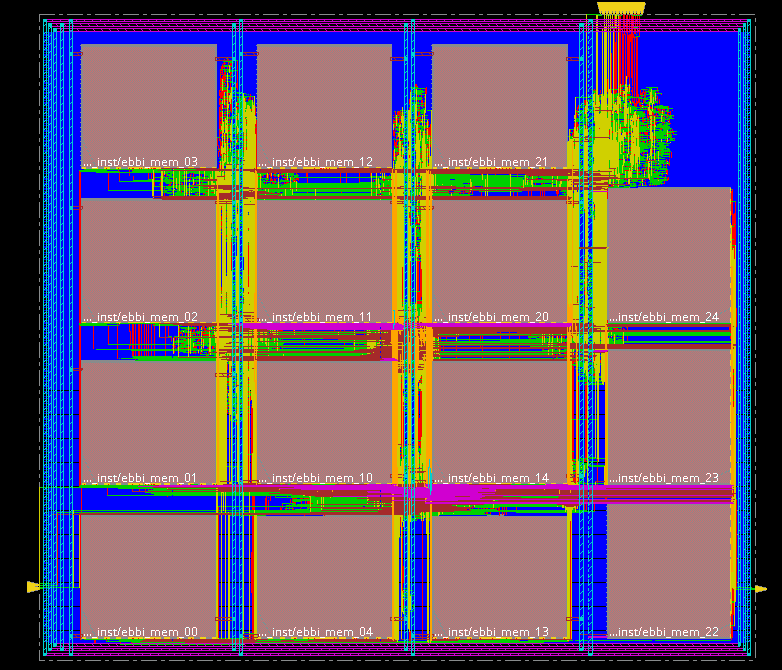}
\caption{ASIC layout of the SNNF in 65nm technology. The 15 SRAM instances implementing the parallel memory banks (labeled \texttt{ebbi\_mem\_00} to \texttt{ebbi\_mem\_04}, \texttt{ebbi\_mem\_10} to \texttt{ebbi\_mem\_14} and \texttt{ebbi\_mem\_20} to \texttt{ebbi\_mem\_24}) preserve the FPGA architecture. The interconnect and logic circuits are also shown.}
\label{fig:layout}
\end{figure}

Fig. \ref{fig:layout} shows the final ASIC layout. The 15 SRAM instances (labeled as \texttt{ebbi\_mem\_00} to \texttt{ebbi\_mem\_04}, \texttt{ebbi\_mem\_10} to \texttt{ebbi\_mem\_14} and \texttt{ebbi\_mem\_20} to \texttt{ebbi\_mem\_24} in the figure) are arranged in a grid, maintaining the parallel memory architecture described in Section \ref{sec:mem}. The colorful routing channels between these modules illustrate the interconnect and logic circuits. Crucially, while the MLPF ASIC architecture requires significant alteration from its FPGA version to avoid the area overhead of instantiating one SRAM per sensor column, the SNNF preserves its efficient architecture. This architectural consistency results in a low, deterministic latency of 9 clock cycles, compared to 33 cycles for the restructured MLPF ASIC.

\begin{table}[!htbp]
\centering
\caption{Performance metrics for ASIC implementation.}
\label{tab:ASIC}
 \begin{tabular}{c|c|c|c|c|c}
\toprule
  \multicolumn{6}{c}{\textbf{65nm ASIC}}\\
 \midrule &\textbf{Logic}&\textbf{SRAM}&\textbf{Power}&\textbf{Freq.}&\textbf{MER}\\ 
 &\textbf{Area} &\textbf{Area} & \textbf{(mW)}&\textbf{(MHz)}&\textbf{(Meps)}\\ 
 \midrule
 \textbf{MLPF} \cite{Rios-Navarro2023} & 0.022 & 4.3 & 40 & 833 & 25.24 \\
 \textbf{SNNF} & 0.056 & 0.51 & 1.48 & 400 & 44.44 \\
\bottomrule
\end{tabular}
\end{table}

Table \ref{tab:ASIC} summarizes the ASIC performance. The total area of SNNF is 0.5656 mm\textsuperscript{2} (Logic: 0.056 mm\textsuperscript{2}, SRAM: 0.5096 mm\textsuperscript{2}), approximately one-eighth of the MLPF's area (4.322 mm\textsuperscript{2}) \cite{Rios-Navarro2023}. Although the SNNF's logic area is larger, the overall area is dominated by the memory. Thus, the saving is primarily driven by the much smaller SRAM footprint. 

Power efficiency is also a key advantage. The energy consumed per event ($E$) is calculated as:
\begin{equation}
E = \frac{P}{f_{\text{sys}}} \times N
\end{equation}
where the dynamic power of the computational block $P = 65.24\,\text{mW}$, the system clock frequency $f_{\text{sys}} = 400\, \text{MHz}$, and the number of clock cycles required to process each event $N = 9$, yielding $E \approx 1.47\,\text{nJ}$ per event. 

At an event rate of $f_{\text{event}} = 1\,\text{Meps}$, the total power consumption $P_{\text{total}} = E \times f_{\text{event}} + P_{\text{leakage}}$, where the leakage power $P_{\text{leakage}} = 0.012\,\text{mW}$. Thus, $P_{\text{total}} \approx 1.47\,\text{nJ} \times 1\,\text{MHz} + 0.012\,\text{mW} = 1.482\,\text{mW}$ ($\approx \mathbf{1.48\,\text{mW}}$). This is over $\mathbf{26\times}$ lower than the $\mathbf{40\,\text{mW}}$ consumed by MLPF. The much smaller SRAM array of our design reduces the static power dissipation substantially.

Furthermore, SNNF achieves a throughput of 44.44 Meps at $400\,\text{MHz}$, outperforming MLPF (25.24 Meps at 833MHz)\cite{Rios-Navarro2023} due to its superior cycle-per-event efficiency (9 clock cycles vs 33 cycles per event for MLPF).

In summary, the ASIC implementation confirms that SNNF achieves a superior power-performance-area trade-off, making it ideal for low-power, near-sensor applications.

\subsection{Comparison With State-of-the-Art Filters}
\label{sec:comparison}

A comprehensive comparison of the proposed SNNF with other hardware-oriented DVS denoisers is presented in Table \ref{tab:filter_comparison}. The filters are evaluated based on filtering accuracy (Driving dataset), memory cell bit-width, memory footprint, and maximum event rate (MER).

\begin{table}[htbp]
\centering
\caption{Comparison with other DVS filters.}
\label{tab:filter_comparison}
\begin{tabularx}{\linewidth}{>{\centering\arraybackslash\hsize=0.8\hsize}X|>{\centering\arraybackslash\hsize=0.5\hsize}X|>{\centering\arraybackslash\hsize=1\hsize}X|>{\centering\arraybackslash\hsize=0.5\hsize}X}
\toprule
\multirow{2}{*}{\textbf{Denoiser}} & \textbf{Driving} & \textbf{Memory} & \textbf{MER} \\
& \textbf{AUC} & \textbf{Size} & \textbf{(Meps)} \\
\midrule
SNNF FPGA & 0.89 & $6 \times W \times H$ & 29 \\
SNNF ASIC & 0.89 & $6 \times W \times H$ & 31.88 \\
MLPF \cite{Rios-Navarro2023} & 0.87 & $18 \times W \times H$ & 23 \\
ONF \cite{Khodamoradi2021} & 0.01 & $64 \times (W + H)$ & 3 \\
HAST \cite{gopalakrishnan2024hast} & 0.84 & $F \times \sqrt{W \times H}$\textsuperscript{b} & 18 \\
HashHeat \cite{Guo2020} & 0.56 & * & 100 \\
\bottomrule
\end{tabularx}
\begin{tablenotes}
    \item a {The bit-width for each memory cell.}
    \item b {$F$ ranges from 0.06-0.22.}
    \item * {Bit width is 128, Memory scaling information not provided.}
\end{tablenotes}

\end{table}



Among the compared filters, SNNF achieves the highest accuracy with an AUC of 0.89 on the driving dataset, outperforming the more complex MLPF (AUC = 0.87) while using a simpler 1-bit memory cell. Regarding memory footprint, SNNF requires $6 \times W \times H$ bits. While this linear scaling is higher than the sub-linear scaling of HAST ($F \times \sqrt{W \times H}$, where $F \in [0.06, 0.22]$) or the constant memory of HashHeat, it enables significantly higher throughput. Specifically, SNNF achieves 29 Meps on FPGA, suprassing HAST (18 Meps) and ONF (3 Meps) while maintaining superior accuracy. Although HashHeat achieves 100 Meps, its AUC of 0.56 is substantially lower, making it unsuitable for high-accuracy applications.

In summary, SNNF offers an excellent trade-off: it delivers state-of-the-art filtering accuracy and high throughput with a manageable, linearly-scaling memory footprint. This makes it exceptionally well-suited for real-time applications where high denoising performance is critical.

\section{Conclusion}
\label{sec:conclusion}

This paper introduced the SNNF, an efficient hardware-friendly filter for DVS event streams. The SNNF employs EBBIs to encode raw event streams into a simplified 2D representation, while an FCSNN classifies events based on extracted spatiotemporal patterns. To achieve high throughput, we proposed a parallel memory bank architecture that facilitates efficient patch extraction.

Through a comprehensive design space exploration, we identified optimal hyperparameters that balance filtering performance with hardware complexity. The proposed architecture was implemented and validated on two FPGA platforms (XC7Z100 and ZU3CG) and in a 65 nm ASIC process. Experimental results demonstrate that the SNNF achieves high denoising accuracy comparable to the best-performing hardware-friendly filters, while providing significant advantages in hardware efficiency. Specifically, it requires substantially lower memory and logic resources, consumes less energy per event, and achieves higher throughput than state-of-the-art alternatives.

These attributes--high accuracy, low resource footprint, energy efficiency, and high throughput--collectively make the SNNF an ideal solution for real-time, near-sensor, and edge-computing applications.

\bibliographystyle{IEEEtran} 
\bibliography{Main}

@ARTICLE{HOST,
  author={Lagorce, Xavier and Orchard, Garrick and Galluppi, Francesco and Shi, Bertram E. and Benosman, Ryad B.},
  journal={IEEE Transactions on Pattern Analysis and Machine Intelligence}, 
  title={HOTS: A Hierarchy of Event-Based Time-Surfaces for Pattern Recognition}, 
  year={2017},
  volume={39},
  number={7},
  pages={1346-1359},
  doi={10.1109/TPAMI.2016.2574707}}

@article{gopalakrishnan2024hast,
  title={Hast: A hardware-efficient spatio-temporal correlation near-sensor noise filter for dynamic vision sensors},
  author={Gopalakrishnan, Pradeep Kumar and Chang, Chip-Hong and Basu, Arindam},
  journal={IEEE Transactions on Circuits and Systems I: Regular Papers},
  year={2024},
  publisher={IEEE}
}

@article{mueggler2017fast,
  title={Fast event-based corner detection},
  author={Mueggler, Elias and Bartolozzi, Chiara and Scaramuzza, Davide},
  year={2017},
  publisher={University of Zurich}
}

@article{zhou2020near,
  title={Near-sensor and in-sensor computing},
  author={Zhou, Feichi and Chai, Yang},
  journal={Nature Electronics},
  volume={3},
  number={11},
  pages={664--671},
  year={2020},
  publisher={Nature Publishing Group UK London}
}

@article{tian2001analysis,
  title={Analysis of temporal noise in CMOS photodiode active pixel sensor},
  author={Tian, Hui and Fowler, Boyd and Gamal, Abbas E},
  journal={IEEE Journal of Solid-State Circuits},
  volume={36},
  number={1},
  pages={92--101},
  year={2001},
  publisher={IEEE}
}

@article{zhu2018ev,
  title={EV-FlowNet: Self-supervised optical flow estimation for event-based cameras},
  author={Zhu, Alex Zihao and Yuan, Liangzhe and Chaney, Kenneth and Daniilidis, Kostas},
  journal={arXiv preprint arXiv:1802.06898},
  year={2018}
}

@inproceedings{devulapally2024multi,
  title={Multi-modal fusion of event and rgb for monocular depth estimation using a unified transformer-based architecture},
  author={Devulapally, Anusha and Khan, Md Fahim Faysal and Advani, Siddharth and Narayanan, Vijaykrishnan},
  booktitle={Proceedings of the IEEE/CVF Conference on Computer Vision and Pattern Recognition},
  pages={2081--2089},
  year={2024}
}

@inproceedings{Delbruck2008,
author = {Tobi Delbruck},
booktitle = {Intl. Symp. on Secure-Life Electronics, Advanced Electronics for Quality Life and Society},
pages = {21-26},
title = {{Frame-free dynamic digital vision}},
year = {2008},
}

@inproceedings{Amir2017,
author = {Arnon Amir and et al},
doi = {10.1109/CVPR.2017.781},
isbn = {978-1-5386-0457-1},
booktitle = {2017 IEEE Conference on Computer Vision and Pattern Recognition (CVPR)},
month = {7},
pages = {7388-7397},
publisher = {IEEE},
title = {{A Low Power, Fully Event-Based Gesture Recognition System}},
volume = {2017-January},
url = {https://ieeexplore.ieee.org/document/8100264/},
year = {2017},
}

@article{Mohan2022,
author = {Vivek Mohan and Deepak Singla and Tarun Pulluri and Andres Ussa and Pradeep Kumar Gopalakrishnan and Pao-Sheng Sun and Bharath Ramesh and Arindam Basu},
doi = {10.1109/JIOT.2022.3178120},
issn = {2327-4662},
journal = {IEEE Internet of Things Journal},
pages = {1-1},
title = {{EBBINNOT: A Hardware-Efficient Hybrid Event-Frame Tracker for Stationary Dynamic Vision Sensors}},
url = {https://ieeexplore.ieee.org/document/9782692/},
year = {2022},
}

@article{Khodamoradi2021,
author = {Khodamoradi, Alireza and Kastner, Ryan},
doi = {10.1109/TETC.2017.2788865},
issn = {21686750},
journal = {IEEE Transactions on Emerging Topics in Computing},
number = {1},
pages = {15--23},
title = {{O(N)-Space Spatiotemporal Filter for Reducing Noise in Neuromorphic Vision Sensors}},
volume = {9},
year = {2021}
}

@article{Bloom1970,
author = {Bloom, Burton H.},
doi = {10.1145/362686.362692},
issn = {15577317},
journal = {Communications of the ACM},
number = {7},
pages = {422--426},
title = {{Space/time trade-offs in hash coding with allowable errors}},
volume = {13},
year = {1970}
}

@article{Guo2022,
author = {Shasha Guo and Tobi Delbruck},
doi = {10.1109/TPAMI.2022.3152999},
isbn = {2021020169},
issn = {0162-8828},
issue = {c},
journal = {IEEE Transactions on Pattern Analysis and Machine Intelligence},
pages = {1-1},
publisher = {IEEE},
title = {{Low Cost and Latency Event Camera Background Activity Denoising}},
volume = {8828},
url = {https://ieeexplore.ieee.org/document/9720086/},
year = {2022},
}

@inproceedings{Acharya2019,
author = {Jyotibdha Acharya and Andres Ussa Caycedo and Vandana Reddy Padala and Rishi Raj Singh Sidhu and Garrick Orchard and Bharath Ramesh and Arindam Basu},
doi = {10.1109/SOCC46988.2019.1570553690},
isbn = {978-1-7281-3483-3},
issn = {21641706},
booktitle = {2019 32nd IEEE International System-on-Chip Conference (SOCC)},
month = {9},
pages = {318-323},
publisher = {IEEE},
title = {{EBBIOT: A Low-complexity Tracking Algorithm for Surveillance in IoVT using Stationary Neuromorphic Vision Sensors}},
volume = {2019-September},
url = {https://ieeexplore.ieee.org/document/9088111/},
year = {2019},
}

@inproceedings{Guo2020,
author = {Guo, Shasha and Kang, Ziyang and Wang, Lei and Li, Shiming and Xu, Weixia},
booktitle = {Proceedings of the Asia and South Pacific Design Automation Conference (ASP-DAC)},
doi = {10.1109/ASP-DAC47756.2020.9045268},
isbn = {9781728141237},
pages = {452--457},
publisher = {IEEE},
title = {{HashHeat: An O(C) Complexity Hashing-based Filter for Dynamic Vision Sensor}},
volume = {2020-Janua},
year = {2020}
}

@article{Guo2021,
author = {Guo, Shasha and Kang, Ziyang and Wang, Lei and Zhang, Limeng and Chen, Xiaofan and Li, Shiming and Xu, Weixia},
doi = {10.1016/j.vlsi.2021.04.006},
issn = {01679260},
journal = {Integration},
number = {2018},
pages = {99--107},
publisher = {Elsevier B.V.},
title = {{HashHeat: A hashing-based spatiotemporal filter for dynamic vision sensor}},
url = {https://doi.org/10.1016/j.vlsi.2021.04.006},
volume = {81},
year = {2021}
}

@article{Delbruck2013,
author = {Tobi Delbruck and Manuel Lang},
doi = {10.3389/fnins.2013.00223},
issn = {1662-453X},
issue = {7 NOV},
journal = {Frontiers in Neuroscience},
pages = {1-9},
title = {{Robotic goalie with 3 ms reaction time at 4\% CPU load using event-based dynamic vision sensor}},
volume = {7},
url = {http://journal.frontiersin.org/article/10.3389/fnins.2013.00223/abstract},
year = {2013},
}

@article{Lichtsteiner2008,
author = {Lichtsteiner, Patrick and Posch, Christoph and Delbruck, Tobi},
journal = {IEEE Journal of Solid-State Circuits},
number = {2},
pages = {566--576},
title = {{A 128x128 120 dB 15us Latency Asynchronous Temporal Contrast Vision Sensor}},
volume = {43},
year = {2008}
}

@inproceedings{Liu2015,
author = {Hongjie Liu and Christian Brandli and Chenghan Li and Shih Chii Liu and Tobi Delbruck},
doi = {10.1109/ISCAS.2015.7168735},
isbn = {9781479983919},
issn = {02714310},
booktitle = {Proceedings - IEEE International Symposium on Circuits and Systems},
pages = {722-725},
publisher = {IEEE},
title = {{Design of a spatiotemporal correlation filter for event-based sensors}},
volume = {2015-July},
year = {2015},
}

@inproceedings{Horowitz2014,
author = {Mark Horowitz},
doi = {10.1109/ISSCC.2014.6757323},
isbn = {9781479909186},
issn = {01936530},
booktitle = {2014 IEEE International Solid-State Circuits Conference Digest of Technical Papers (ISSCC)}, 
pages = {10-14},
title = {{Computing's Energy Problem (and what we can do about it)}},
volume = {57},
year = {2014},
}

@inproceedings{Hu2021,
title = {{v2e: From Video Frames to Realistic DVS Events}},
booktitle = "2021 {IEEE/CVF} Conference on Computer Vision and Pattern Recognition Workshops ({CVPRW})",
author = "Hu, Y and Liu, S C and Delbruck, T",
publisher = "IEEE",
year = 2021,
url = "http://arxiv.org/abs/2006.07722"
}

@article{Gallego2022,
author = {Guillermo Gallego and Tobi Delbrück and Garrick Orchard and Chiara Bartolozzi and Brian Taba and Andrea Censi and Stefan Leutenegger and Andrew Davison and Jörg Conradt and Kostas Daniilidis and Davide Scaramuzza},
doi = {10.1109/tpami.2020.3008413},
issn = {23318422},
journal = {IEEE Trans. Pattern Anal. Mach. Intell.},
volume = {44},
number = {1},
pages = {154-180},
title = {{Event-based vision: A survey}},
year = {2022},
}

@article{Benosman2012,
author = {Ryad Benosman and Sio Hoi Ieng and Charles Clercq and Chiara Bartolozzi and Mandyam Srinivasan},
doi = {10.1016/j.neunet.2011.11.001},
issn = {08936080},
journal = {Neural Networks},
pages = {32-37},
pmid = {22154354},
publisher = {Elsevier Ltd},
title = {{Asynchronous frameless event-based optical flow}},
volume = {27},
url = {http://dx.doi.org/10.1016/j.neunet.2011.11.001},
year = {2012},
}

@article{Ussa2023,
author = {A. Ussa et. al},
doi = {10.1109/TNNLS.2023.3243679},
journal = {IEEE Transactions on Neural Networks and Learning Systems},
publisher = {IEEE},
title = {{A Hybrid Neuromorphic Object Tracking and Classification Framework for Real-Time Systems}},
volume = {Early Access},
year = {2023},
}

@article{Benosman2014,
author = {Ryad Benosman and Charles Clercq and Xavier Lagorce and Sio Hoi Ieng and Chiara Bartolozzi},
doi = {10.1109/TNNLS.2013.2273537},
issn = {2162237X},
issue = {2},
journal = {IEEE Transactions on Neural Networks and Learning Systems},
pages = {407-417},
pmid = {24807038},
publisher = {IEEE},
title = {{Event-Based Visual Flow}},
volume = {25},
year = {2014},
}

@inproceedings{Rebecq2019,
author = {Rebecq, Henri and Ranftl, Rene and Koltun, Vladlen and Scaramuzza, Davide},
title = {{Events-To-Video: Bringing Modern Computer Vision to Event Cameras}},
booktitle = {Proceedings of the IEEE/CVF Conference on Computer Vision and Pattern Recognition (CVPR)},
month = {June},
year = {2019}
}

@article{Tapiador2020,
author = {Ricardo Tapiador-Morales and Jean Matthieu Maro and Angel Jimenez-Fernandez and Gabriel Jimenez-Moreno and Ryad Benosman and Alejandro Linares-Barranco},
doi = {10.3390/s20123404},
issn = {14248220},
issue = {12},
journal = {Sensors (Switzerland)},
month = {6},
pages = {1-16},
pmid = {32560238},
publisher = {MDPI AG},
title = {{Event-based gesture recognition through a hierarchy of time-surfaces for FPGA}},
volume = {20},
year = {2020},
}

@inproceedings{Maqueda2018,
author = {Maqueda, Ana I. and Loquercio, Antonio and Gallego, Guillermo and García, Narciso and Scaramuzza, Davide},
booktitle = {2018 IEEE/CVF Conference on Computer Vision and Pattern Recognition}, 
title = {{Event-Based Vision Meets Deep Learning on Steering Prediction for Self-Driving Cars}}, 
year = {2018},
volume = {},
number = {},
pages = {5419-5427},
doi = {10.1109/CVPR.2018.00568}}

@article{Rebecq2021,
author = {Rebecq, Henri and Ranftl, René and Koltun and Vladlen and Scaramuzza, Davide},  journal = {IEEE Transactions on Pattern Analysis and Machine Intelligence},
title = {{High Speed and High Dynamic Range Video with an Event Camera}},   
year = {2021},
volume = {43},
number = {6},
pages = {1964-1980},  doi = {10.1109/TPAMI.2019.2963386}}

@inproceedings{Gehrig2019,
author = {Daniel Gehrig and Antonio Loquercio and Konstantinos Derpanis and Davide Scaramuzza},
doi = {10.1109/ICCV.2019.00573},
isbn = {978-1-7281-4803-8},
issn = {15505499},
booktitle = {2019 IEEE/CVF International Conference on Computer Vision (ICCV)},
month = {10},
pages = {5632-5642},
publisher = {IEEE},
title = {{End-to-End Learning of Representations for Asynchronous Event-Based Data}},
volume = {2019-October},
url = {https://ieeexplore.ieee.org/document/9009469/},
year = {2019},
}

@inproceedings{Zhu2018,
author = {Zhu, Alex Zihao and Yuan, Liangzhe and Chaney, Kenneth and Daniilidis, Kostas},
booktitle = {2019 IEEE/CVF Conference on Computer Vision and Pattern Recognition (CVPR)}, 
title = {{Unsupervised Event-Based Learning of Optical Flow, Depth, and Egomotion}}, 
year = {2019},
volume = {},
number = {},
pages = {989-997},
doi = {10.1109/CVPR.2019.00108}}

@ARTICLE{Nie2021,
author = {Nie, Kaiming and Shi, Xiaopei and Cheng, Silu and Gao, Zhiyuan and Xu, Jiangtao},
journal = {IEEE Transactions on Circuits and Systems for Video Technology}, 
title = {{High Frame Rate Video Reconstruction and Deblurring Based on Dynamic and Active Pixel Vision Image Sensor}}, 
year = {2021},
volume = {31},
number = {8},
pages = {2938-2952},
doi = {10.1109/TCSVT.2020.3034148}}

@ARTICLE{Ramesh2021,
author = {Ramesh, Bharath and Zhang, Shihao and Yang, Hong and Ussa, Andres and Ong, Matthew and Orchard, Garrick and Xiang, Cheng},
journal = {IEEE Transactions on Circuits and Systems for Video Technology}, 
title = {{e-TLD: Event-Based Framework for Dynamic Object Tracking}}, 
year = {2021},
volume = {31},
number = {10},
pages = {3996-4006},
doi = {10.1109/TCSVT.2020.3044287}}

@INPROCEEDINGS {Rios-Navarro2023,
author = {A. Rios-Navarro and S. Guo and G. Abarajithan and K. Vijayakumar and A. Linares-Barranco and T. Aarrestad and R. Kastner and T. Delbruck},
booktitle = {2023 IEEE/CVF Conference on Computer Vision and Pattern Recognition Workshops (CVPRW)},
title = {{Within-Camera Multilayer Perceptron DVS Denoising}},
year = {2023},
volume = {},
issn = {},
pages = {3933-3942},
doi = {10.1109/CVPRW59228.2023.00409},
url = {https://doi.ieeecomputersociety.org/10.1109/CVPRW59228.2023.00409},
publisher = {IEEE Computer Society},
address = {Los Alamitos, CA, USA},
month = {jun}
}

@ARTICLE{Eshraghian2023,
author={Eshraghian, Jason K. and Ward, Max and Neftci, Emre O. and Wang, Xinxin and Lenz, Gregor and Dwivedi, Girish and Bennamoun, Mohammed and Jeong, Doo Seok and Lu, Wei D.},
journal={Proceedings of the IEEE}, 
title={Training Spiking Neural Networks Using Lessons From Deep Learning}, 
year={2023},
volume={111},
number={9},
pages={1016-1054},
doi={10.1109/JPROC.2023.3308088}}

@ARTICLE{TrueNorth2015,
author={Akopyan, Filipp and Sawada, Jun and Cassidy, Andrew and Alvarez-Icaza, Rodrigo and Arthur, John and Merolla, Paul and Imam, Nabil and Nakamura, Yutaka and Datta, Pallab and Nam, Gi-Joon and Taba, Brian and Beakes, Michael and Brezzo, Bernard and Kuang, Jente B. and Manohar, Rajit and Risk, William P. and Jackson, Bryan and Modha, Dharmendra S.},
journal={IEEE Transactions on Computer-Aided Design of Integrated Circuits and Systems}, 
title={TrueNorth: Design and Tool Flow of a 65 mW 1 Million Neuron Programmable Neurosynaptic Chip}, 
year={2015},
volume={34},
number={10},
pages={1537-1557},
doi={10.1109/TCAD.2015.2474396}}

@ARTICLE{Loihi2018,
author={Davies, Mike and Srinivasa, Narayan and Lin, Tsung-Han and Chinya, Gautham and Cao, Yongqiang and Choday, Sri Harsha and Dimou, Georgios and Joshi, Prasad and Imam, Nabil and Jain, Shweta and Liao, Yuyun and Lin, Chit-Kwan and Lines, Andrew and Liu, Ruokun and Mathaikutty, Deepak and McCoy, Steven and Paul, Arnab and Tse, Jonathan and Venkataramanan, Guruguhanathan and Weng, Yi-Hsin and Wild, Andreas and Yang, Yoonseok and Wang, Hong},
journal={IEEE Micro}, 
title={Loihi: A Neuromorphic Manycore Processor with On-Chip Learning}, 
year={2018},
volume={38},
number={1},
pages={82-99},
doi={10.1109/MM.2018.112130359}}

@article{haessig2018spiking,
  title={Spiking optical flow for event-based sensors using ibm's truenorth neurosynaptic system},
  author={Haessig, Germain and Cassidy, Andrew and Alvarez, Rodrigo and Benosman, Ryad and Orchard, Garrick},
  journal={IEEE transactions on biomedical circuits and systems},
  volume={12},
  number={4},
  pages={860--870},
  year={2018},
  publisher={IEEE}
}

@inproceedings{Brosch2016,
author = {Brosch, Tobias and Neumann, Heiko},
title = {Event-based optical ﬂow on neuromorphic hardware},
year = {2016},
isbn = {9781631901003},
publisher = {ICST (Institute for Computer Sciences, Social-Informatics and Telecommunications Engineering)},
address = {Brussels, BEL},
url = {https://doi.org/10.4108/eai.3-12-2015.2262447},
doi = {10.4108/eai.3-12-2015.2262447},
booktitle = {Proceedings of the 9th EAI International Conference on Bio-Inspired Information and Communications Technologies (Formerly BIONETICS)},
pages = {551–558},
numpages = {8},
location = {New York City, United States},
series = {BICT'15}
}

@InProceedings{Dikov2017,
author="Dikov, Georgi
and Firouzi, Mohsen
and R{\"o}hrbein, Florian
and Conradt, J{\"o}rg
and Richter, Christoph",
title="Spiking Cooperative Stereo-Matching at 2 ms Latency with Neuromorphic Hardware",
booktitle="Biomimetic and Biohybrid Systems",
year="2017",
publisher="Springer International Publishing",
address="Cham",
pages="119--137",
isbn="978-3-319-63537-8"
}

@inproceedings{andreopoulos2018low,
  title={A Low Power, High Throughput, Fully Event-Based Stereo System},
  author={Andreopoulos, Alexander and Kashyap, Hirak J and Nayak, Tapan K and Amir, Arnon and Flickner, Myron D},
  booktitle={2018 IEEE/CVF Conference on Computer Vision and Pattern Recognition (CVPR)},
  pages={7532--7542},
  year={2018},
  organization={IEEE Computer Society}
}

@misc{Glover2019,
title={ATIS + SpiNNaker: a Fully Event-based Visual Tracking Demonstration}, 
author={Arren Glover and Alan B. Stokes and Steve Furber and Chiara Bartolozzi},
year={2019},
eprint={1912.01320},
archivePrefix={arXiv},
primaryClass={cs.CV},
url={https://arxiv.org/abs/1912.01320}, 
}

\newcommand*{\OnlineSM}{}

\ifdefined\OnlineSM
\onecolumn


\addtocontents{toc}{\protect\setcounter{tocdepth}{2}}

\section*{\LARGE SNNF: An SNN-based Near-Sensor Noise Filter\\for Dynamic Vision Sensors - SUPPLEMENTARY MATERIAL.}

\Large 

\renewcommand\contentsname{}
\tableofcontents

\newpage
\phantomsection
\addcontentsline{toc}{section}{Section A - CSNN performance}
\begin{center}
\subsection*{\LARGE Section A -CSNN performance}
\end{center}

\noindent Fig. \ref{fig:csnn_ebbi}, \ref{fig:csnn_kernals} shows the schematic
diagram and the results of the sweeps for the CSNN, with the
results indicating that CSNN does not offer any performance
advantage over the 1-layer FCSNN.
It can also be seen that the energy and hardware resource
requirements of both the 2-layer network and CSNN are
much higher than that of the 1-layer FCSNN as shown in
Section C. Therefore, we did not
choose to adopt these architectures or study them further, in
this work.

\begin{figure} [h!]
\centering
\begin{minipage}{0.45\textwidth}
\centering
\includegraphics[width=\textwidth]{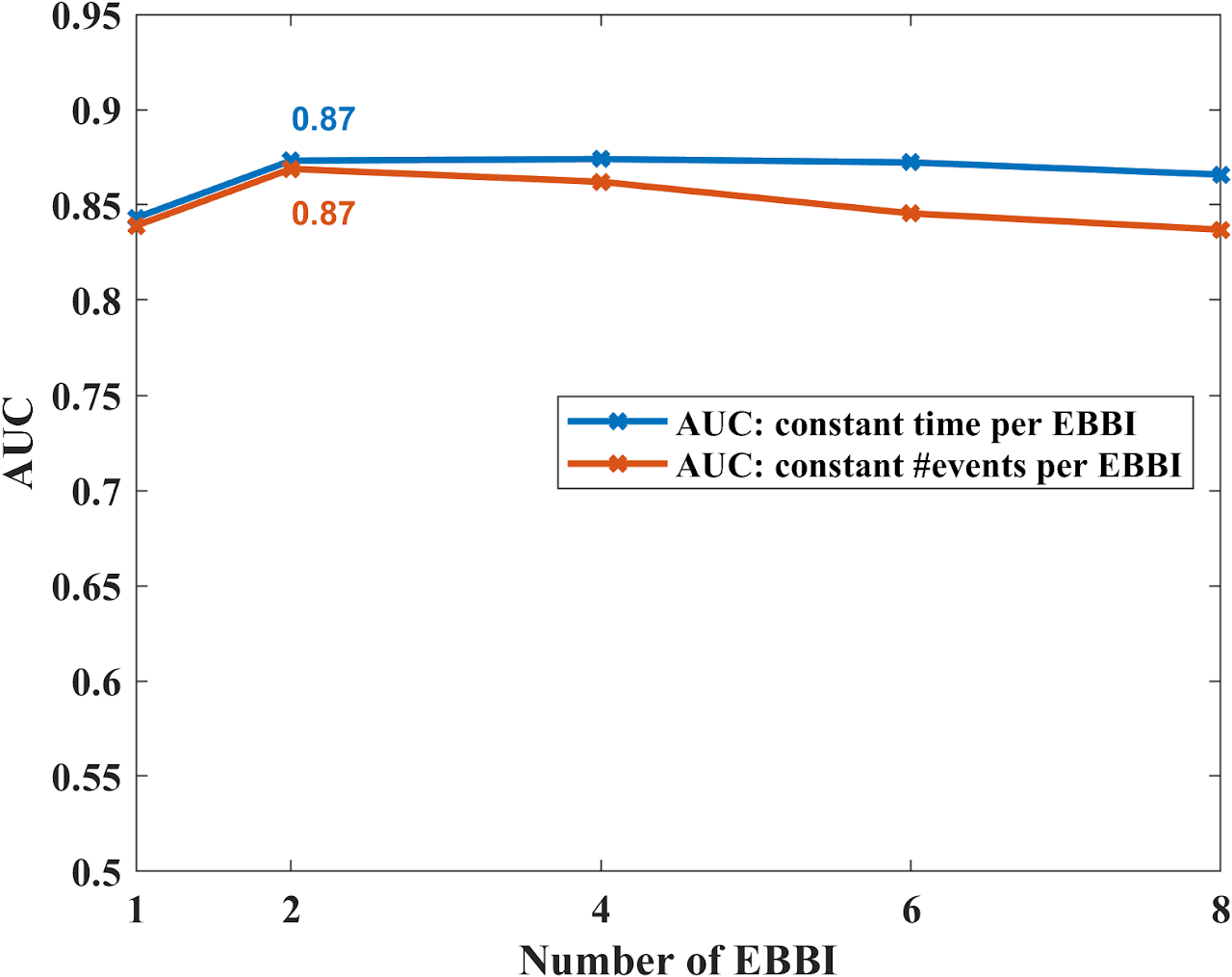}
\caption{CSNN: AUC vs. number of EBBI pairs ($N_{\text{EBBI}})$}
\label{fig:csnn_ebbi}
\end{minipage}\hfill
\begin{minipage}{0.45\textwidth}
\centering
\includegraphics[width=\textwidth]{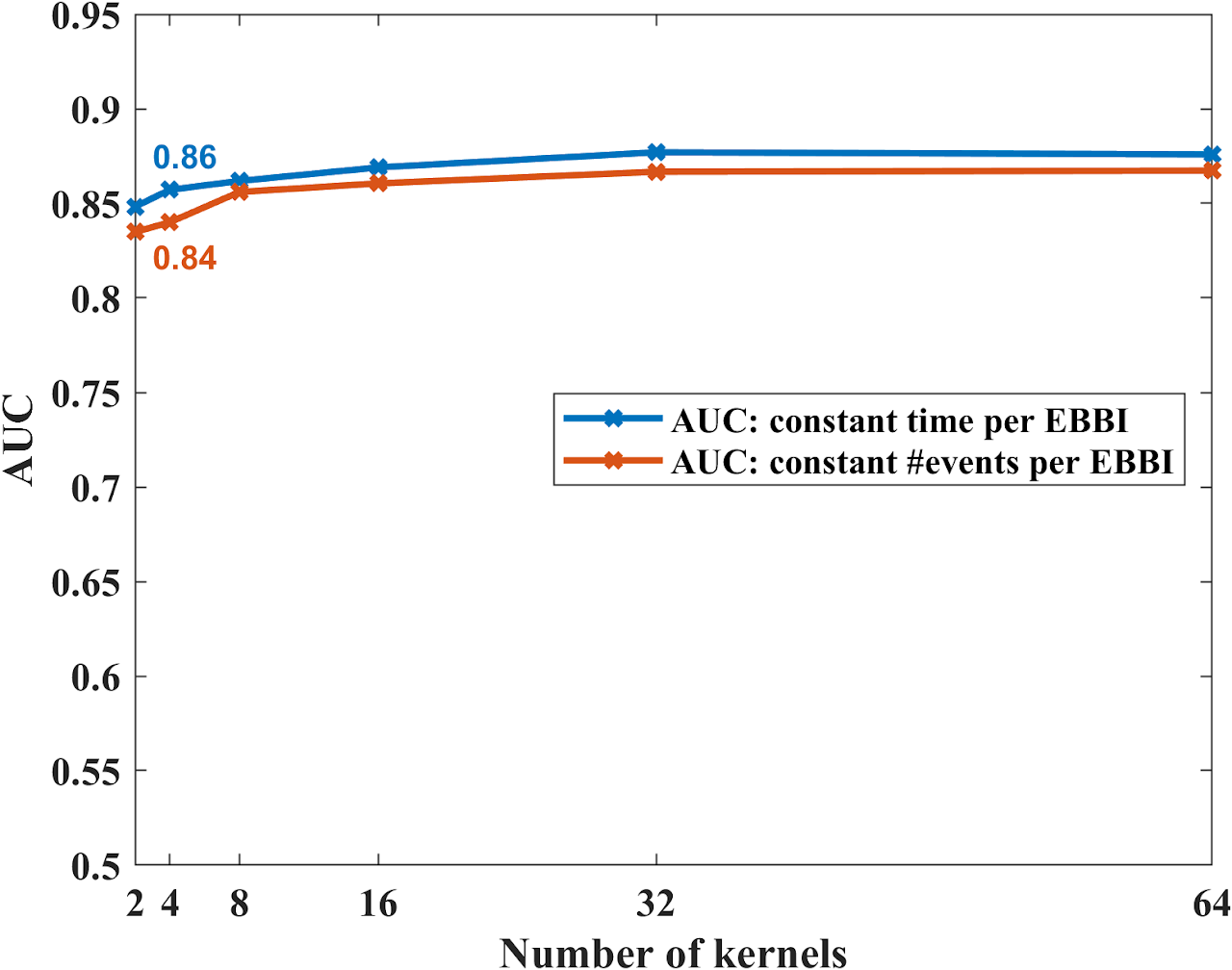}
\caption{CSNN: AUC vs. number of kernals}
\label{fig:csnn_kernals}
\end{minipage}

\end{figure}

\newpage
\phantomsection
\addcontentsline{toc}{section}{Section B - Results for Hotel-bar dataset}
\begin{center}
\subsection*{\LARGE Section B - Results for Hotel-bar dataset }
\end{center}

\noindent Hotel bar is a typical surveillance operation dataset. This dataset is created from aggressively de-noised outputs from recordings---with a DAVIS346 event camera---of scenes from a hotel lobby. The video is approximately $12$ minutes long, with a resolution of $346$ $\times$ $260$ pixels. We used a 1-minute segment of the video with about $1.58$ M events, for running our simulations. As with the \textit{driving} dataset, the \textit{leak noise} and \textit{shot noise} files were added to the noise-free video files and input to the filters, as for the dataset described earlier.

When sweeping the Hotel Bar dataset, we observe from Fig. \ref{fig:sweep_ebbi} that under the constant number of events method, EBBI still achieves its highest AUC with $N_{\text{EBBI}} = 2$ (4 EBBIs total). However, under the constant time method, EBBI reaches its peak AUC only when $N_{\text{EBBI}} =8$, which incurs higher resource consumption.

Similarly, by sweeping both methods, we observe that each achieves the same maximum AUC of 0.96. Fig. \ref{fig:sweep_train} shows a detailed hyperparameter sweep on the hotel bar dataset. However, the constant number of events method reaches this peak AUC when $N_e = 20,000$, whereas the constant time method requires 100 ms to attain the same performance. Therefore, for the Hotel Bar dataset, the constant number of events approach is more efficient and preferable.

Similarly, a sweep conducted on the hotel-bar dataset indicates that the constant number of events method is marginally superior to the constant time method. However, for this dataset, the maximum AUC of 0.958 was achieved at \textit{$\tau$}=100ms, as illustrated in the figure.

\begin{figure} [h!]
\centering
\begin{minipage}{0.45\textwidth}
\centering
\includegraphics[width=\textwidth]{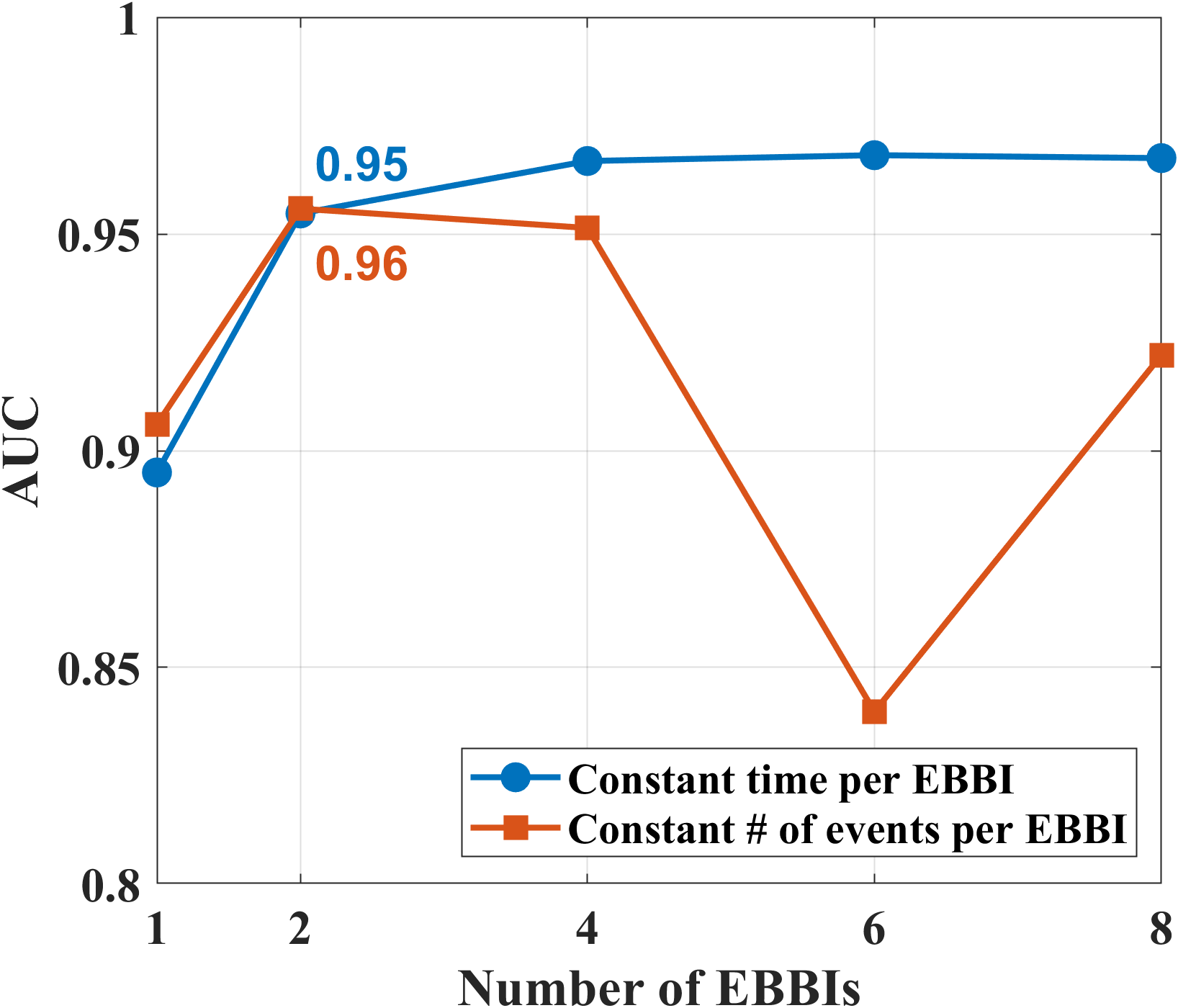}
\caption{AUC vs. number of EBBI pairs ($N_{\text{EBBI}})$}
\label{fig:sweep_ebbi}
\end{minipage}\hfill
\begin{minipage}{0.45\textwidth}
\centering
\includegraphics[width=\textwidth]{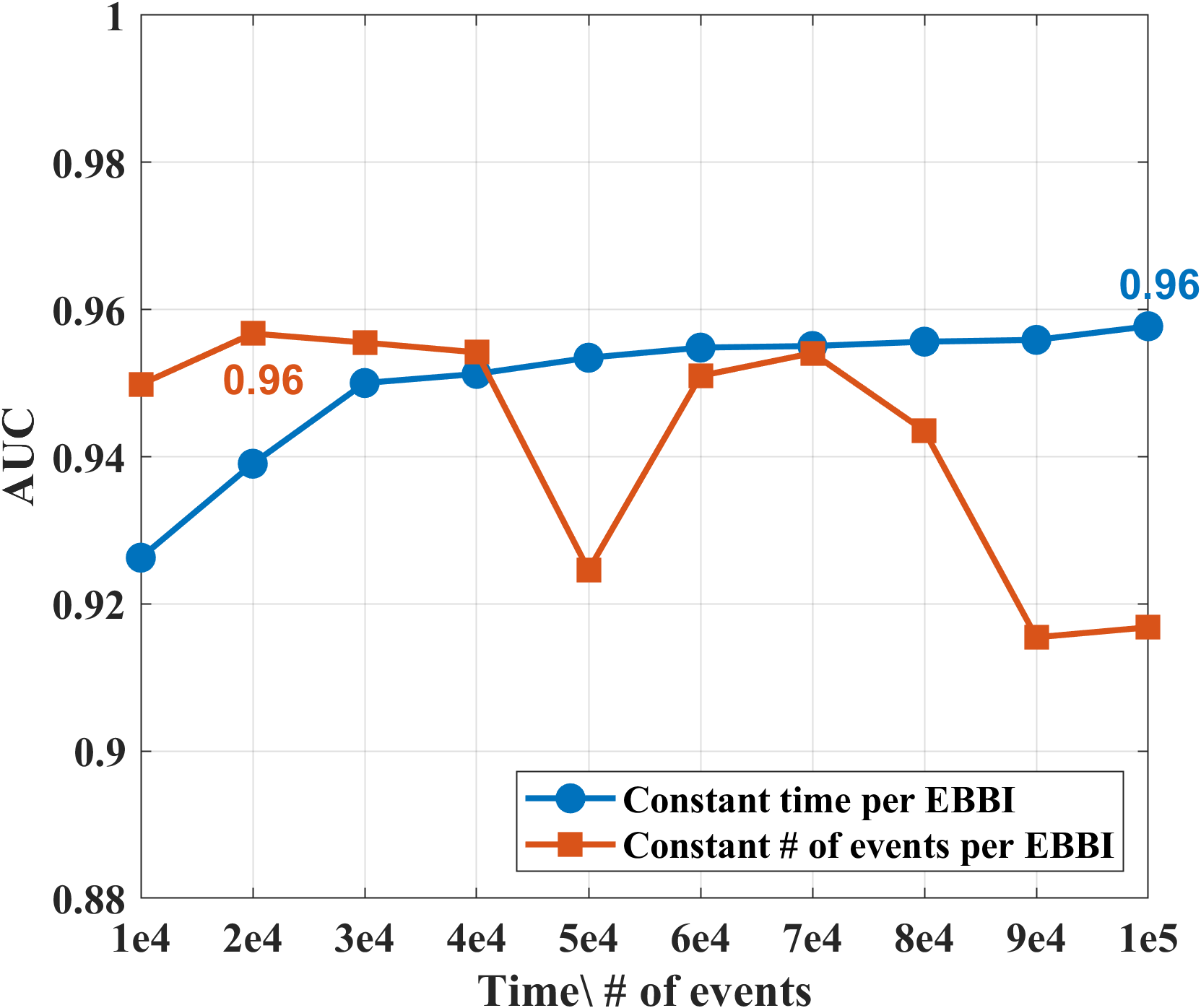}
\caption{AUC vs. the EBBI creation hyperparameters: time interval $T_e$ and event count $N_e$}
\label{fig:sweep_train}
\end{minipage}

\end{figure}

Fig. \ref{fig:fpr_vs_tpr_hotel_bar} shows the FPR-TPR curves of the filters when
simulated using a 6-second segment of the hotel-bar dataset
from DND21 mixed with shot noise. As with the driving
dataset, SNNIF and MLPF present the highest values for AUC.

\newpage
\begin{figure}[htb]
\centering
\captionsetup[subfigure]{font=scriptsize, labelfont=scriptsize}
\includegraphics[width=0.5 \textwidth]{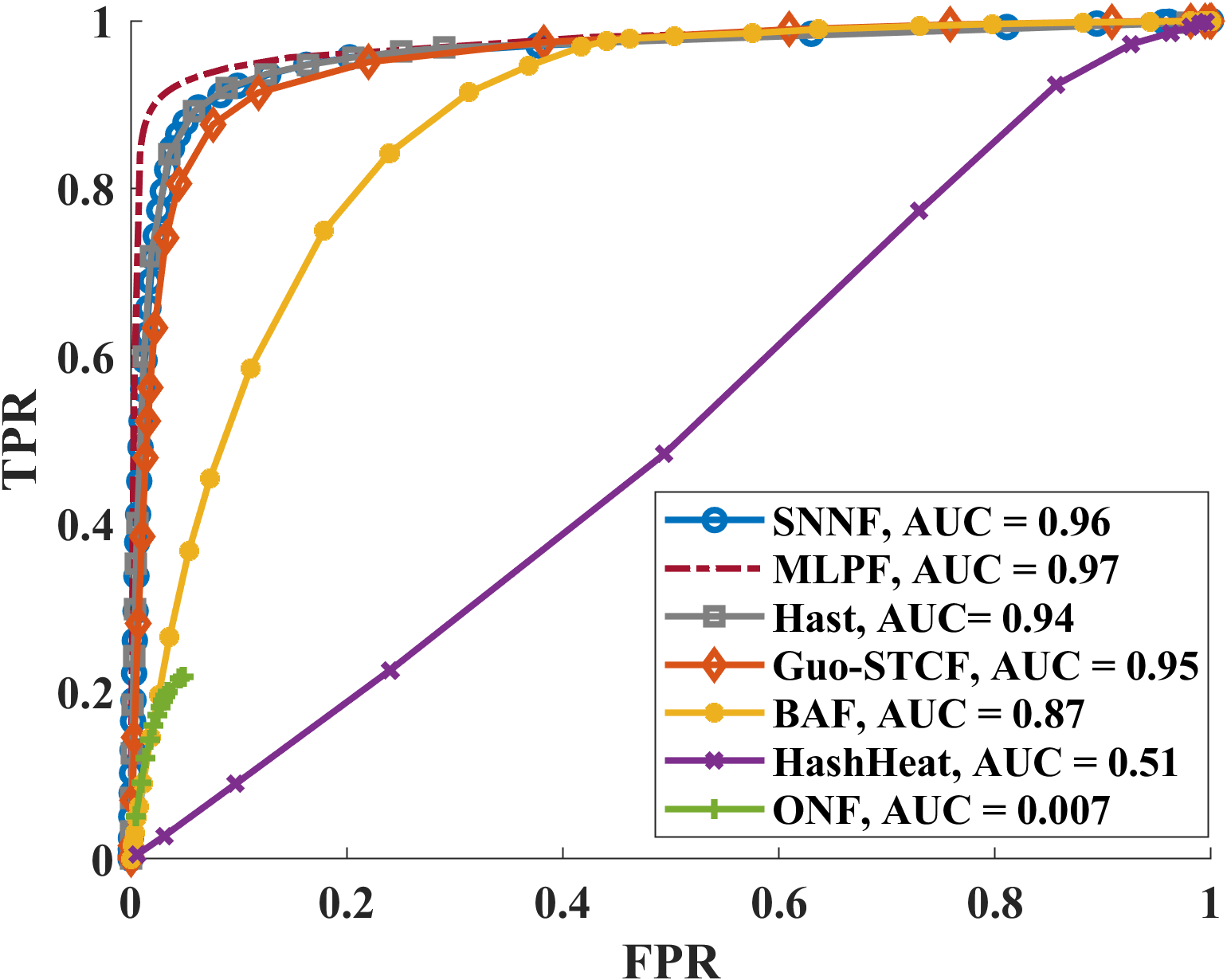}
\caption{FPR-TPR ROC curve and AUC of \FNAME{} vs. other filters for \textit{hotel-bar} data mixed with \textit{shot noise}}
\label{fig:fpr_vs_tpr_hotel_bar}
\end{figure}

\newpage
\phantomsection
\addcontentsline{toc}{section}{Section C - Memory requirement and  energy (pJ) per event calculation}
\begin{center}
\subsection*{\LARGE Section C - Memory requirement and  energy (pJ) per event calculation}
\end{center}

\vspace{1em}

\begin{table}[!htbp]
\centering
\caption{Energy estimates of filters vs. sensor sizes (pJ/event).}
\label{tab:energy}
\begin{tabular}{lcccc}
\toprule
Filter & 240$\times$180 & 346$\times$260 & 640$\times$480 & 1280$\times$960 \\
\midrule
BAF/STCF & 149  & 215  & 523  & 1829 \\
MLPF     & 417  & 531  & 1062 & 3312 \\
HashHeat & 21   & 21   & 21   & 21   \\
ONF      & 67   & 72   & 85   & 114  \\
SNNF     & 208  & 219  & 266  & 315  \\
\bottomrule
\end{tabular}
\end{table}

\begin{table}[!htbp]
\centering
\caption{Memory requirement of filters vs. sensor sizes (KB).}
\label{tab:memory}
\begin{tabular}{lcccc}
\toprule
Filter & 240$\times$180 & 346$\times$260 & 640$\times$480 & 1280$\times$960 \\
\midrule
BAF/STCF       & 169  & 351  & 1200 & 4800 \\
MLPF           & 96   & 198  & 676  & 2701 \\
HashHeat       & 32   & 32   & 32   & 32   \\
ONF            & 6.56 & 9.46 & 17.5 & 35   \\
SNNF           & 33   & 67   & 227  & 902  \\
\bottomrule
\end{tabular}
\end{table}

\noindent Although HashHeat's memory and energy consumption remain constant, its performance is low compared to most other filters, even for low-resolution sensors such as 346 $\times$ 260, compared to most other filters, as shown by the FPR versus TPR curves in Fig. \ref{fig:fpr_vs_tpr} (main text).

Please see the ancillary file: \textit{Section\_C\_SNNF\_vs\_other\_filters.xlsx} for the details of calculation of energy (pJ) per event and Memory requirement.
\newpage
\phantomsection
\addcontentsline{toc}{section}{Section D - FPGA implementation details}

\begin{center}
\subsection*{\LARGE Section D - FPGA implementation details}
\vspace{2em}
\subsection*{\LARGE Section D.1 - FPGA top level diagram}
\end{center} 

\begin{figure} [htbp]
\centering 
\includegraphics[width=0.95\columnwidth]{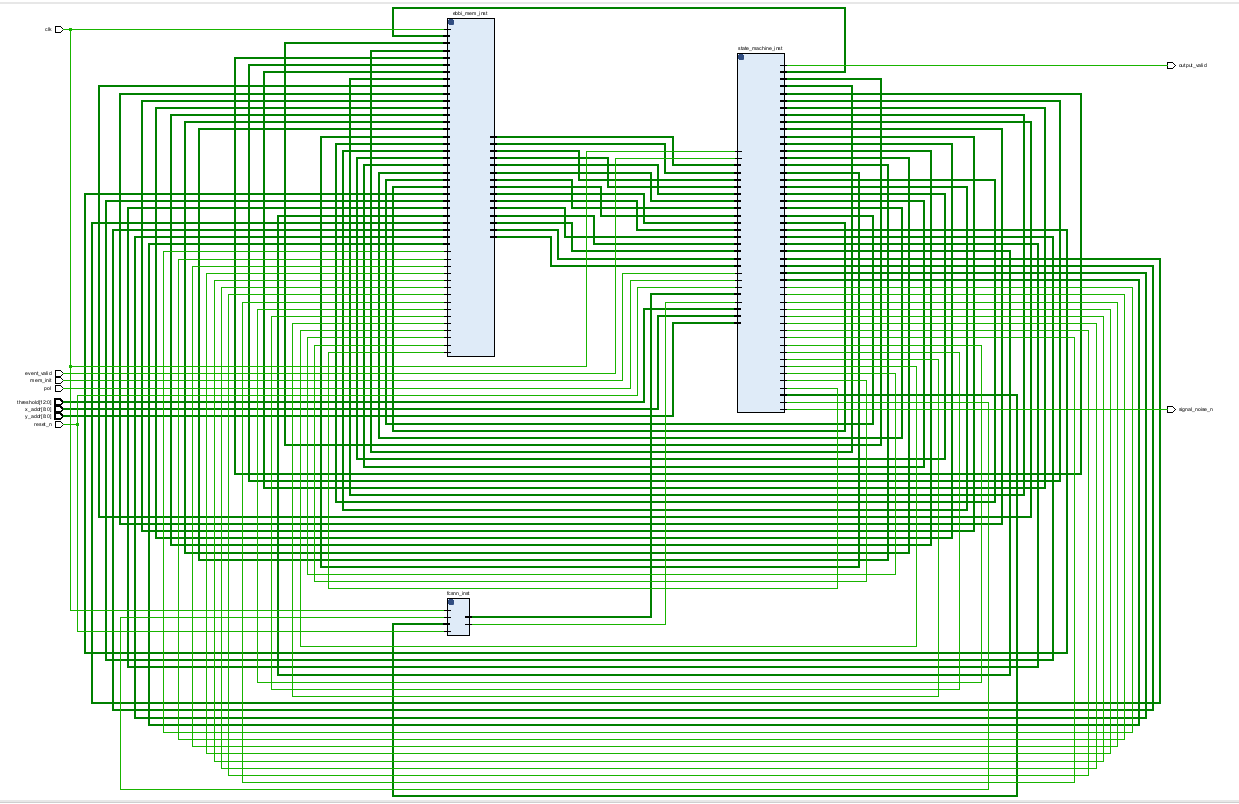}
\setlength{\belowcaptionskip}{-10pt}
 \caption{FPGA top-level diagram}
\label{fig:new_plot}
\end{figure}

\vspace{2em}

\noindent The system consists of three parts: EBBI\_MEM, State machine and FCSNN network.


\newpage
\begin{center}
\subsection*{\LARGE Section D.2 - FPGA Power Estimation}
\end{center} 

\begin{figure} [htbp]
\centering 
\includegraphics[width=0.95\columnwidth]{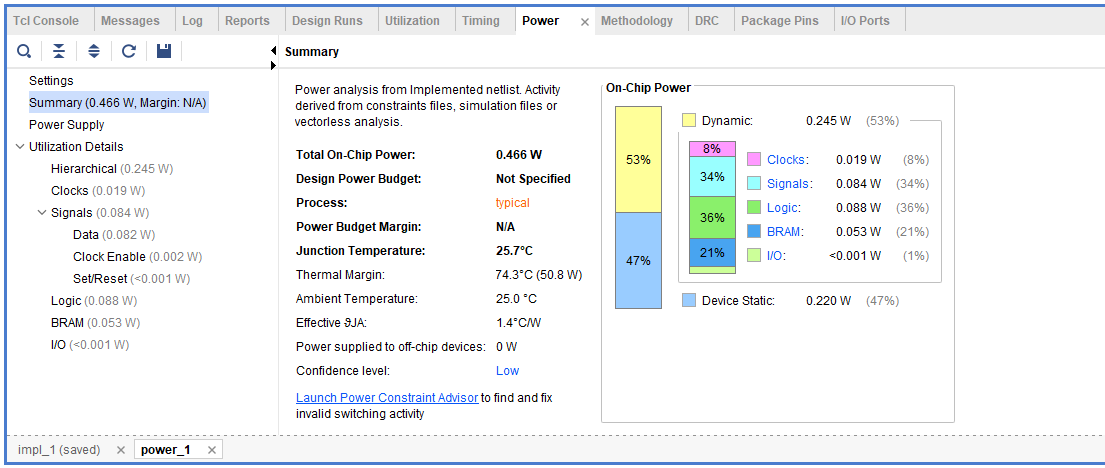}
\setlength{\belowcaptionskip}{-10pt}
 \caption{FPGA Power estimation results. FPGA Device used: XC7Z100.}
\label{fig:fpga_power}
\end{figure}

\begin{figure} [htbp]
\centering 
\includegraphics[width=0.95\columnwidth]{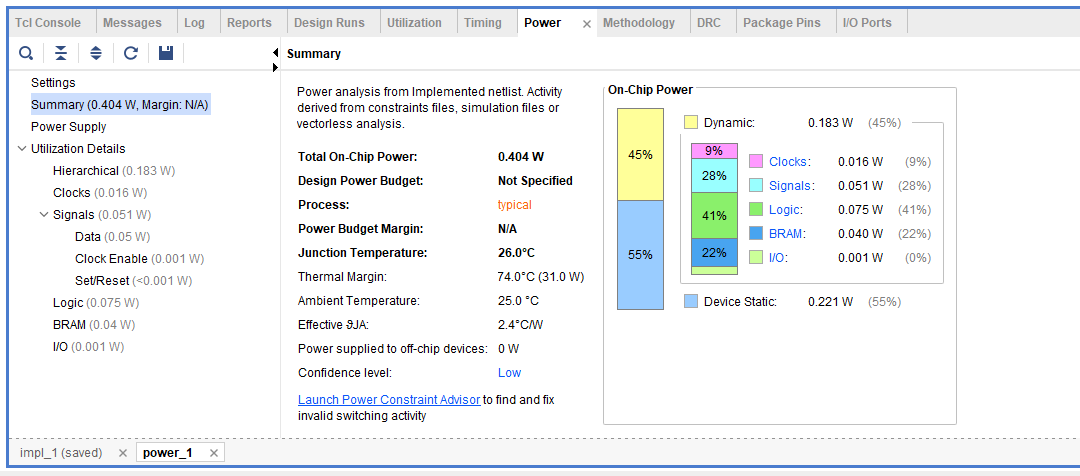}
\setlength{\belowcaptionskip}{-10pt}
 \caption{FPGA Power estimation results. FPGA Device used: ZU3CG.}
\label{fig:fpga_power_zu3cg}
\end{figure}

\newpage
\begin{center}
\subsection*{\LARGE Section D.3 - FPGA Resource Comparison}
\end{center}

\noindent The specific FPGA resource utilization is as follows:
\begin{table}[htbp]
\centering
\large
\caption{\large XC7Z100: Resource Utilization}
\label{tab:hash_comparison}
\begin{tabular}{|c|c|c|c|}
\hline
\multicolumn{1}{|c|}{\textbf{Resource}} & \multicolumn{1}{c|}{\textbf{Utilization}} & \multicolumn{1}{c|}{\textbf{Available}}& \multicolumn{1}{c|}{\textbf{Utilization \%}} \\ \hline
LUT & 7037 & 277400 & 2.54 \\ \hline
FF & 2101 & 554800 & 0.38 \\ \hline
BRAM & 22.5 & 755 & 2.98 \\ \hline
\end{tabular}
\end{table}
\begin{table}[htbp]
\centering
\large
\caption{\large ZU3CG: Resource Utilization}
\label{tab:hash_comparison}
\begin{tabular}{|c|c|c|c|}
\hline
\multicolumn{1}{|c|}{\textbf{Resource}} & \multicolumn{1}{c|}{\textbf{Utilization}} & \multicolumn{1}{c|}{\textbf{Available}}& \multicolumn{1}{c|}{\textbf{Utilization \%}} \\ \hline
LUT & 6676 & 70560 & 9.46 \\ \hline
FF & 2104 & 141120 & 1.49 \\ \hline
BRAM & 22.5 & 216 & 10.42 \\ \hline
\end{tabular}
\end{table}
\fi

\end{document}